\documentclass[11pt,a4paper]{article}
\usepackage[hyperref]{acl2021}
\usepackage{times}
\usepackage{latexsym}

\usepackage{array}
\newcolumntype{C}[1]{>{\centering\let\newline\\\arraybackslash\hspace{0pt}}m{#1}}

\usepackage{graphicx}

\usepackage{amsmath}

\usepackage{multirow}

\usepackage{floatrow}
\newfloatcommand{capbtabbox}{table}[][\FBwidth]
\usepackage{subfig}
\usepackage{caption}

\DeclareMathOperator*{\argmin}{arg\,min}

\usepackage{microtype}

\aclfinalcopy 

\title{Inspecting the concept knowledge graph \\ encoded by modern language models}

\author{Carlos Aspillaga$^1$, \ Marcelo Mendoza$^2$, \ Alvaro Soto$^1$ \\
  $^1$ Computer Science Department, Pontificia Universidad Católica de Chile \\
  $^2$ Department of Informatics, Universidad Técnica Federico Santa María, Chile \\
  \texttt{cjaspill@uc.cl,mmendoza@inf.utfsm.cl,asoto@ing.puc.cl} \\}

\date{}

\begin{document}
\maketitle
\begin{abstract}
The field of natural language understanding has experienced exponential progress in the last few years, with impressive results in several tasks. This success has motivated researchers to study the underlying knowledge encoded by these models. Despite this, attempts to understand their semantic capabilities have not been successful, often leading to non-conclusive, or contradictory conclusions among different works. Via a probing classifier, we extract the underlying knowledge graph of nine of the most influential language models of the last years, including word embeddings, text generators, and context encoders. This probe is based on concept relatedness, grounded on WordNet. Our results reveal that all the models encode this knowledge, but suffer from several inaccuracies. Furthermore, we show that the different architectures and training strategies lead to different model biases. We conduct a systematic evaluation to discover specific factors that explain why some concepts are challenging. We hope our insights will motivate the development of models that capture concepts more precisely.
\end{abstract}

\section{Introduction}

Natural language processing (NLP) encompasses a wide variety of applications such as summarization \citep{Kovaleva:19}, information retrieval \citep{Zhan:20}, and machine translation \citep{Tang:18a}, among others. Currently, the use of pre-trained language models has become the \textit{de facto} starting point to tackle most of these tasks. The usual pipeline consists of finetuning a pre-trained language model by using a discriminative learning objective to adapt the model to the requirements of each task. As key ingredients, these models are pre-trained using massive amounts of unlabeled data that can include millions of documents and billions of parameters. Massive data and parameters are supplemented with a suitable learning architecture, resulting in a highly powerful but also complex model whose internal operation is hard to analyze.

The success of pre-trained language models has driven the interest to understand the mechanisms they use to solve NLP tasks. As an example, in the case of BERT~\citep{Devlin:19}, one of the most popular pre-trained models based on the Transformer~\citep{Vaswani:17}, several studies have attempted to access the knowledge encoded in its layers and attention heads~\citep{Tenney:19a,Devlin:19,Hewitt:19}. In particular, \citet{Jawahar:19} shows that BERT can solve tasks at a syntactic level by using Transformer blocks to encode a soft hierarchy of features at different levels of abstraction. Similarly, \citet{Hewitt:19} show 
that BERT is capable of encoding structural information from text. In particular, using a structural probe, they show that syntax trees are embedded in a linear transformation of the encodings of BERT. 

In general, previous efforts have provided strong evidence indicating that current pre-trained language models encode complex syntactic rules. However, relevant evidence about their abilities to capture semantic information remains still elusive. As an example, \citet{Si:19} attempts to locate the encoding of semantic information as part of the top layers of Transformer architectures finding contradictory evidence. Similarly, \citet{Kovaleva:19} focuses on studying knowledge encoded by self-attention weights. Their results provide evidence for over-parameterization but not about language understanding capabilities. 

In this work, we study to what extent pre-trained language models encode semantic information. As a key source of semantic knowledge, we analyze their ability to encode the concept relations embedded in the conceptual taxonomy of WordNet\footnote{WordNet is a human-generated graph, where each one of its 117000 nodes (also called synsets) represent a concept. In this work, we use hyponymy relations, representing if a concept is a subclass of another.}~\citep{Miller:95}. Understanding, organizing, and correctly using concepts is one of the most remarkable capabilities of human intelligence~\citep{Lake:EtAl:MachineHuman:2017}. Therefore, quantifying the ability that a pre-trained language model can exhibit to encode the conceptual organization behind WordNet is highly valuable. This knowledge may provide useful insights into the inner mechanisms that these models use to encode semantic information.
Furthermore, identifying what they find difficult can provide relevant insights into how to improve them.

Unlike most previous works, we do not focus on a particular model but target a large list of the most popular pre-trained language models. In this sense, one of our goals is to provide a comparative analysis of the benefits of different approaches. Following \citet{Hewitt:19}, we study semantic performance by defining a probing classifier based on concept relatedness according to WordNet. Using this tool, we analyze the different models, enlightening how and where semantic knowledge is encoded. Furthermore, we explore how these models encode suitable information to recreate the structure of WordNet. Among our main results, we show that the different pre-training strategies and architectures lead to different model biases. In particular, we show that contextualized word embeddings, such as BERT, encode high-level concepts and hierarchical relationships among them, creating a taxonomy. This finding corroborates previous work results \citep{Reif:19} that claim that BERT vectors store sub-spaces that correspond with semantic knowledge. Our study also shows evidence about the limitations of current pre-trained language models, demonstrating that they have difficulties to encode specific concepts. For example, all the models struggle with concepts related to ``taxonomical groups''. Our results also reveal that models have distinctive patterns regarding where in the architecture they encode the semantic information. These patterns are dependant on architecture and not on model sizes.

\section{Study methodology}
\label{section:methodology}

\begin{figure*}[ht!]
    \centering
    \includegraphics[width=11cm]{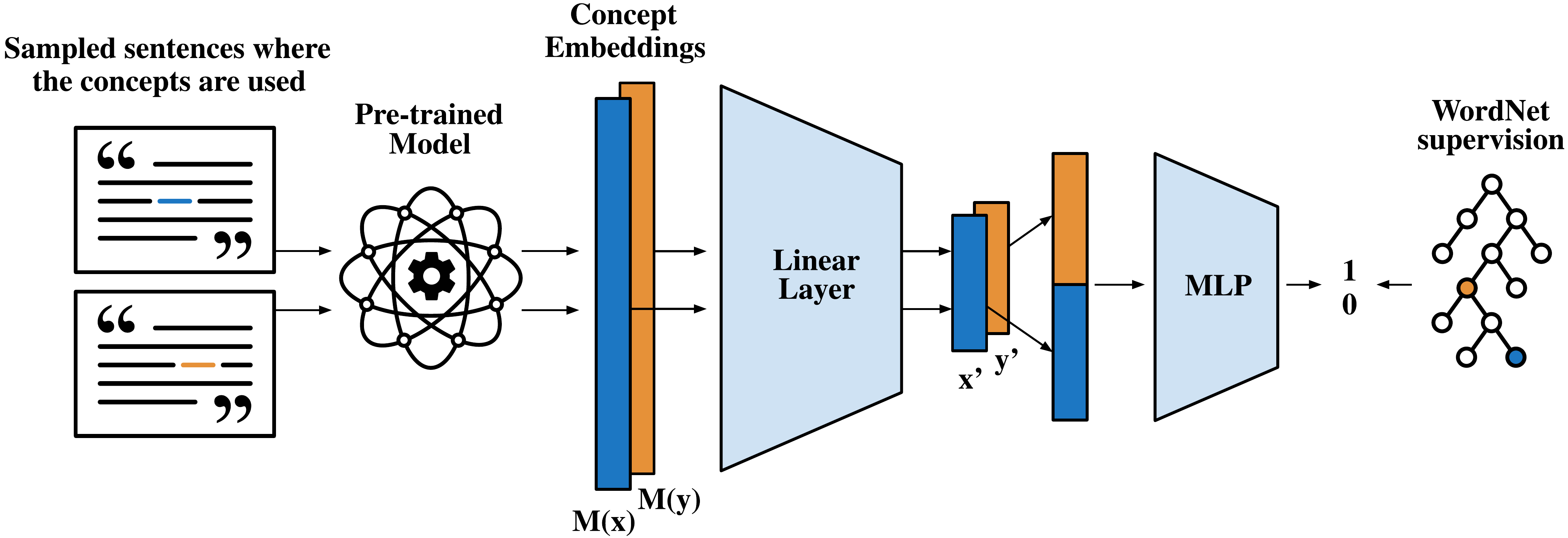}
    \caption{Inputs to the edge probing classifier correspond to the model embeddings $M(x)$ and $M(y)$ of concepts $x$ and $y$, respectively. $M(x)$ and $M(y)$ are projected into a common lower dimensionality space using a linear layer. The resulting embeddings $x'$ and $y'$ are concatenated and fed into a Multi-Layer Perceptron that is in charge of predicting if the concept pair is related or not.}
    \label{fig:edge}
\end{figure*}

Probing methods consist of using the representation of a frozen pre-trained model to address a particular task. If the probing classifier succeeds in this setting but fails using an alternative model, it means that the source model encodes the knowledge needed to solve the task. Furthermore, the classifier's performance can be used to measure how well the model captures this knowledge \citep{conneau:2018:cram}. We use a probing method at the semantic level applying it to the nine models presented in Section \ref{section:selected-models}. 
Our study sheds light on whether the models encode relevant knowledge to predict concept relatedness in Wordnet.

To study how accurately the models encode semantic information, we measure correctness in predicted relations among concepts at two levels: (a) pair-wise-level by studying performance across sampled pairs of related or unrelated concepts, and (b) graph-level by using pair-wise predictions to reconstruct the actual graph. We describe both approaches in Sections \ref{probe-classifier-methodology} and \ref{reconstruction-methodology}, respectively. 


\subsection{WordNet splits and sampling}
\label{section:experimental-setup}
We partitioned the available WordNet synsets at 70/15/15 for training, validation and test sets respectively. Our experimental setup ensures no overlap in concepts among these sets. As an example, if the concept related to ``house" fell in the training set, then all its lemmas are considered in this partition (e.g. ``home", ``residence", etc.), and neither this concept nor those lemmas will be present in the validation or test sets. 
Our sampling setup also balances the number of times each concept acts as hypernym or as a hyponym in the relation, whenever possible. Thus the benefit of learning whether a word is a ``prototypical hypernym", as pointed out by \citet{Levy:15}, is close to zero. Further details are available in Appendix \ref{app:imp-1}.

\subsection{Word embedding models}
\label{section:selected-models}
This study considers the most influential language models from recent years. We consider the essential approaches of three model families: non contextualized word embeddings (NCE), contextualized word embeddings (CE), and generative language models (GLM). We consider Word2Vec \citep{Mikolov:13} and GloVe \citep{Pennington:14} for the first family of approaches. For the CE family, we consider ELMo \citep{Peters:18}, which is implemented on a bidirectional LSTM architecture, XLNet \citep{Yang:19}, and BERT \citep{Devlin:19} and its extensions ALBERT \citep{Lan:20} and RoBERTa \citep{Liu:19b}, all of them based on the Transformer architecture. GPT-2 \citep{Radford:19} and T5 \citep{Raffel:20} are included in the study to incorporate approaches based on generative language models. 

For models in the CE and GLM families, the embedding is extracted after running the model on a sentence where the concept is used in context. Then we discard the context and keep only the first token that correspond to the specific mention of the concept. Finally we concatenate the hidden states of every layer of the model, for the selected token.


\subsection{Semantic probing classifier}
\label{probe-classifier-methodology}




\begin{figure*}[ht!]
\begin{floatrow}
\ffigbox[7.7cm]{%
   \includegraphics[height=147pt]{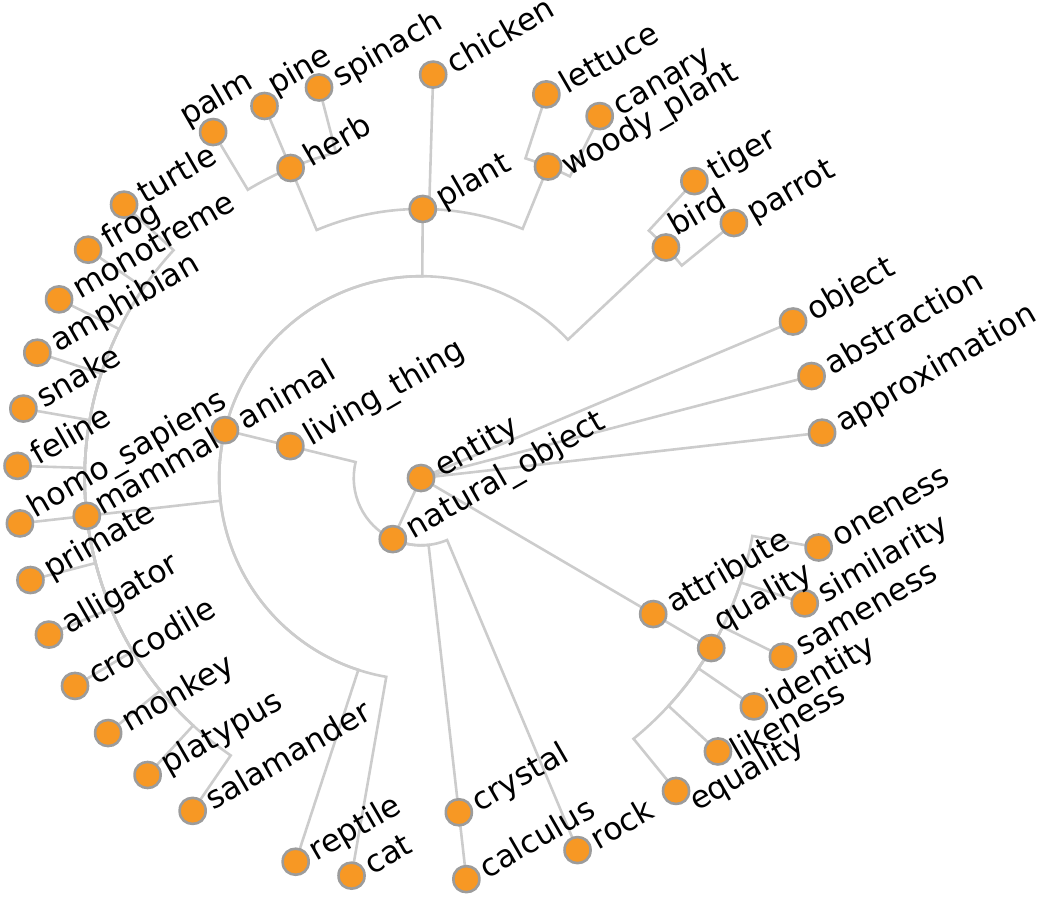}
}{%
  \caption{A reconstructed graph using BERT-large. Visual inspection reveals that the models capture key categories but fail to map fine-grained relations.}%
  \label{fig:knowledgeGraph}
}
\capbtabbox[7.7cm]
{%
  \begin{tabular}{llccc}
        \hline
        \multicolumn{1}{c}{\multirow{2}{*}{\textbf{\footnotesize{Family}}}} & \multicolumn{1}{c}{\multirow{2}{*}{\textbf{\footnotesize{Model}}}} & \multicolumn{3}{c}{\textbf{\footnotesize{Tree Edit Dist.}}} \\ \cline{3-5} 
        \multicolumn{1}{c}{} & \multicolumn{1}{c}{} & \textbf{\footnotesize{TIM}} & \textbf{\footnotesize{MCM}} & \textbf{\footnotesize{Avg.}} \\ \hline
        \multirow{2}{*}{\footnotesize{NCE}} & \footnotesize{Word2Vec} & \footnotesize{59} & \footnotesize{59} & \textbf{\footnotesize{59}} \\
         & \footnotesize{GloVe-42B} & \footnotesize{56} & \footnotesize{60} & \textbf{\footnotesize{58}} \\ \hline
        \multirow{2}{*}{\footnotesize{GLM}} & \footnotesize{GPT-2} & \footnotesize{53} & \footnotesize{57} & \textbf{\footnotesize{55}} \\
         & \footnotesize{T5} & \footnotesize{58} & \footnotesize{55} & \textbf{\footnotesize{56}} \\ \hline
        \multirow{5}{*}{\footnotesize{CE}} & \footnotesize{ELMo} & \footnotesize{52} & \footnotesize{55} & \textbf{\footnotesize{53}} \\
         & \footnotesize{BERT} & \footnotesize{49} & \footnotesize{48} & \textbf{\footnotesize{49}} \\
         & \footnotesize{RoBERTa} & \footnotesize{56} & \footnotesize{54} & \textbf{\footnotesize{55}} \\
         & \footnotesize{XLNet} & \footnotesize{52} & \footnotesize{48} & \textbf{\footnotesize{50}} \\
         & \footnotesize{ALBERT} & \footnotesize{53} & \footnotesize{50} & \textbf{\footnotesize{51}} \\ \hline
        \end{tabular}
}
{%
  \caption{Tree Edit Distance against the ground truth graph (large models used). We display both strategies for estimating $d_e$ along with their average score.}
  \label{table:tree-edit-distances}
}
\end{floatrow}
\end{figure*}

We define an edge probing classifier that learns to identify if two concepts are semantically related. To create the probing classifier, we retrieve all the glosses from the Princeton WordNet Gloss Corpus\footnote{\url{https://wordnetcode.princeton.edu/glosstag.shtml}}. This dataset provides WordNet's synsets gloss sentences with annotations identifying occurrences of concepts within different sentence contexts. The annotations provide a mapping of the used words to their corresponding WordNet node. We sample hypernym pairs A, B. Then, from an unrelated section of the taxonomy, we randomly sample a third synset C, taking care that C is not related to either A or B. Then, $\langle A, B, C \rangle$ forms a triplet that allows us to create six testing edges for our classifier. To train the probing classifier, we define a labeled edge $\{x, y, L\}$, with $x$ and $y$ synsets in $\{A, B, C\}$, $x \neq y$. $L \in \{0, 1\}$ is the target of the edge. If $y$ is direct or indirect parent of $x$, $L = 1$, while $L = 0$ in other case. For each synset $x$, $y$, we sample one of its sentences $S(x)$, $S(y)$ from the dataset. Let $M$ be a model. If $M$ belongs to the NCE family, $x$ and $y$ are encoded by $M(x)$ and $M(y)$, respectively. If $M$ belongs to the CE or GLM families, then $x$ and $y$ are encoded by the corresponding token of $M(S(x))$ and $M(S(y))$, respectively. 


To facilitate the evaluation of embeddings of different sizes, we first project each concept's encodings $x$ and $y$ into a low dimensionality space using a linear layer (see Figure \ref{fig:edge}). These vectors, denoted as $x'$ and $y'$, are concatenated and fed into a Multi-Layer Perceptron (MLP) classifier. The linear layer and the MLP are the only trainable parameters of our setting, as we use the source model weights without any finetuning.
Throughout all the experiments we used an MLP classifier with a single hidden layer of 384 hidden units.

We use this MLP to learn the structural relation between concept pairs, providing the test with a mechanism that allows the embeddings to be combined in a non-linear way. Tests based on linear transformations such as the one proposed by \citet{Hewitt:19} did not allow us to recover the WordNet structure. This indicates that the sub-spaces where the language models encode semantics are not linear. 
The fact that syntactic information is linearly available suggests that syntax trees might be a critical intermediate result for the language modeling task. In contrast, semantic information emerges as an indirect consequence of accurate language modeling. Still, it might not constitute information that the model relies on for NLP tasks, as postulated by \citet{Ravichander:2020}.

To discard the possibility of the MLP being memorizing properties of words and thus giving an undeserved credit to the analyzed models, we generated alternative training and validation sets with random word embeddings of the same size as the real ones. During training and inference, these vectors were kept frozen. These tests showed around $50\%$ accuracy in the binary classification task, indicating that the MLP cannot do better than chance in that scenario. Thus, if in a later experiment the same MLP succeeds at the task, the merit can be attributed to the input embedding itself. This result is consistent with the fact that our experimental setup ensures no overlap in concepts among training, development, and testing sets.

\begin{table*}[h!]
\caption{Results obtained using the edge probing classifier. We study the performance in many model variants, considering small and large versions of several models. Results are grouped by method families.}
\label{table:models-performance}
\begin{center}
\begin{tabular}{p{1cm}lcccc}
 &  & \textbf{\footnotesize{Emb. Size}} & \textbf{\footnotesize{Best}} & \textbf{\footnotesize{F1-score}} & \textbf{\footnotesize{F1-score}} \\ [-0.5ex]
\textbf{\footnotesize{Family}} & \textbf{\footnotesize{Model}} & \footnotesize{\textbf{All/Best Layer}} & \textbf{\footnotesize{Layer}} & \footnotesize{\textbf{All Layers}} & \footnotesize{\textbf{Best Layer}} \\ 
\hline \\ [-2.5ex]
\multirow{1.7}{*}{\footnotesize{NCE}} & \footnotesize{Word2Vec} \scriptsize{\citep{Mikolov:13}} & \footnotesize{300 / -} & \footnotesize{-} & \footnotesize{.7683} \scriptsize{${\pm}$ .0135} & \footnotesize{-} \\ [-0.3ex]
 & \footnotesize{GloVe-42B}  \scriptsize{\citep{Pennington:14}} & \footnotesize{300} / - & \footnotesize{-} & \footnotesize{.7877} \scriptsize{${\pm}$ .0084} & \footnotesize{-} \\ [-0.3ex]
\hline \\ [-2.5ex]
\multirow{3.6}{*}{\footnotesize{GLM}} & \footnotesize{GPT-2} \scriptsize{\citep{Radford:19}} & \footnotesize{9984 / 768} & \footnotesize{6} & \footnotesize{.7862} \scriptsize{${\pm}$ .0132} & \footnotesize{.7921} \scriptsize{${\pm}$ .0108} \\ [-0.3ex]
 & \footnotesize{T5-small} \scriptsize{\citep{Raffel:20}} & \footnotesize{7168 / 512} & \footnotesize{4} & \footnotesize{.8156} \scriptsize{${\pm}$ .0098} & \footnotesize{.8199} \scriptsize{${\pm}$ .0081} \\ [-0.3ex]
 \cline{2-6} \\ [-2.5ex]
 & \footnotesize{GPT2-xl} \scriptsize{\citep{Radford:19}} & \footnotesize{78400 / 1600} & \footnotesize{13} & \footnotesize{.7946} \scriptsize{${\pm}$ .0151} & \footnotesize{.8029} \scriptsize{${\pm}$ .0118} \\ [-0.3ex]
 & \footnotesize{T5-large} \scriptsize{\citep{Raffel:20}} & \footnotesize{51200 / 1024} & \footnotesize{17} & \footnotesize{.8148} \scriptsize{${\pm}$ .0119} & \footnotesize{.8331} \scriptsize{${\pm}$ .0102} \\ [-0.3ex]
\hline \\ [-2.5ex]
\multirow{9}{*}{\footnotesize{CE}} & \footnotesize{ELMo-small} \scriptsize{\citep{Peters:18}} & \footnotesize{768 / 256} & \footnotesize{2} & \footnotesize{.7986} \scriptsize{${\pm}$ .0126} & \footnotesize{.7880} \scriptsize{${\pm}$ .0119} \\ [-0.3ex]
 & \footnotesize{BERT-base} \scriptsize{\citep{Devlin:19}} & \footnotesize{9984 / 768} & \footnotesize{10} & \footnotesize{.8240} \scriptsize{${\pm}$ .0123} & \footnotesize{.8185} \scriptsize{${\pm}$ .0104} \\ [-0.3ex]
 & \footnotesize{RoBERTa-base} \scriptsize{\citep{Liu:19b}} & \footnotesize{9984 / 768} & \footnotesize{5} & \footnotesize{.8392} \scriptsize{${\pm}$ .0100} & \footnotesize{.8266} \scriptsize{${\pm}$ .0083} \\ [-0.3ex]
 & \footnotesize{XLNet-base} \scriptsize{\citep{Yang:19}} & \footnotesize{9984 / 768} & \footnotesize{4} & \footnotesize{.8306} \scriptsize{${\pm}$ .0113} & \footnotesize{.8293} \scriptsize{${\pm}$ .0116} \\ [-0.3ex]
 & \footnotesize{ALBERT-base} \scriptsize{\citep{Lan:20}} & \footnotesize{9984 / 768} & \footnotesize{12} & \footnotesize{.8184} \scriptsize{${\pm}$ .0222} & \footnotesize{.8073} \scriptsize{${\pm}$ .0102} \\ [-0.3ex]
 \cline{2-6} \\ [-2.5ex]
 & \footnotesize{ELMo-large} \scriptsize{\citep{Peters:18}} & \footnotesize{3072 / 1024} & \footnotesize{2} & \footnotesize{.8311} \scriptsize{${\pm}$ .0090} & \footnotesize{.8330} \scriptsize{${\pm}$ .0083} \\ [-0.3ex]
 & \footnotesize{BERT-large} \scriptsize{\citep{Devlin:19}} & \footnotesize{25600 / 1024} & \footnotesize{14} & \footnotesize{.8178} \scriptsize{${\pm}$ .0152} & \footnotesize{.8185} \scriptsize{${\pm}$ .0113} \\ [-0.3ex]
 & \footnotesize{RoBERTa-large} \scriptsize{\citep{Liu:19b}} & \footnotesize{25600 / 1024} & \footnotesize{13} & \footnotesize{.8219} \scriptsize{${\pm}$ .0159} & \footnotesize{.8314} \scriptsize{${\pm}$ .0082} \\ [-0.3ex]
 & \footnotesize{XLNet-large} \scriptsize{\citep{Yang:19}} & \footnotesize{25600 / 1024} & \footnotesize{6} & \footnotesize{.8211} \scriptsize{${\pm}$ .0142} & \footnotesize{.8244} \scriptsize{${\pm}$ .0080} \\ [-0.3ex]
 & \footnotesize{ALBERT-xxlarge} \scriptsize{\citep{Lan:20}} & \footnotesize{53248 / 4096} & \footnotesize{4} & \footnotesize{.8233} \scriptsize{${\pm}$ .0107} & \footnotesize{.8194} \scriptsize{${\pm}$ .0097} \\ [-0.3ex]
\hline
\end{tabular}
\end{center}
\end{table*}




\subsection{Reconstructing the structure of a knowledge graph}
\label{reconstruction-methodology}

The probe classifier predicts if a pair of concepts $\langle u,v \rangle$ form a valid $\langle\text{parent},\text{child} \rangle$ relation according to WordNet, where $h_{\langle u,v \rangle} \in [0,1]$ denotes the corresponding classifier output. It is important to note that valid $\langle\text{parent},\text{child} \rangle$ relations include direct relations (e.g.  $\langle\text{dog},\text{poodle} \rangle$), and transitive relations (e.g. $\langle\text{animal},\text{poodle} \rangle$), and that the order of the items matters.  

To reconstruct the underlying knowledge graph, for each valid $\langle\text{parent},\text{child} \rangle$ relation given by $h_{\langle u,v \rangle} > \text{threshold}$, we need an estimation of how close are the nodes in the graph. We do this by introducing the concept of ``parent closeness'' between a parent node $u$ and a child node $v$, denoted by $d_e(u,v)$. We propose two alternative scores to estimate $d_e$:

\textbf{i) Model Confidence Metric (MCM)}: All the models considered in this study capture close relations more precisely than distant relations (supporting evidence can be found in Appendix \ref{section:aditional-factors}). This means that a concept like \textit{poodle} will be matched with its direct parent node \textit{dog} with higher confidence than with a more distant parent node (e.g. \textit{animal}). Thus, we can define $d_e(u,v) = 1 - h_{\langle u,v \rangle}$.

\textbf{ii) Transitive Intersections Metric (TIM)}: We explore a metric grounded directly in the tree structure of a knowledge graph. Note that nodes \textit{u} and \textit{v} that form a parent-child relation have some transitive connections in common. Specifically, all descendants of \textit{v} are also descendants of \textit{u}, and all the ancestors of \textit{u} are also ancestors of \textit{v}. Then, the closer the link between \textit{u} and \textit{v} in the graph, the bigger the intersection. Accordingly, for each edge $e = \langle u,v \rangle$, we define $d_e(u,v)$ as:
\begin{equation} \label{eq:tim}
- \bigg( \sum_{j \in N \setminus \{u,v\} }{h_{\langle u,j \rangle} h_{\langle v,j \rangle} + h_{\langle j,u \rangle} h_{\langle j,v \rangle}} \bigg) * h_{\langle u,v \rangle},
\end{equation}
where the first term of the sum accounts for the similarity within the descendants of nodes $u$ and $v$, and the second term accounts for the similarity within the ancestors of nodes $u$ and $v$. The term $h_{\langle u,v \rangle}$ at the right-hand side accounts for the edge direction, and $N$ denotes the set of nodes (concepts).

A strategy to find a tree that comprises each node's closest parents is the minimum-spanning-arborescence (MSA) of the graph defined using $d_e$. The MSA is analogous to the minimum-spanning-tree (MST) objective used by \citet{Hewitt:19}, but for directed graphs. The formulation of the MSA optimization problem applied to our proposal is provided in the Appendix \ref{msa-formulation}.

\section{How accurate is this knowledge?}
\label{section:how-accurate}

\subsection{Semantic edge probing classifier results}
\label{section:prediction-performance-comparison}




Table \ref{table:models-performance} shows the results obtained using the edge probing classifier. Results show that regardless of model sizes, performance is homogeneous within each family of models. Additionally, results show that NCE and GLM methods obtain a worse performance when all the layers are used than those achieved by CE methods. 
When single layers are used, GLM shows improved performance, suggesting that these models capture semantics earlier in the architecture,  keeping their last layers for generative-specific purposes. In contrast, CE models degrade or maintain their performance when single layers are used.

Note that Table \ref{table:models-performance} shows pair-wise metrics not graph metrics. As we are dealing with graphs, predicted edges are built upon related edges. Thus, drifts in small regions of the graph may cause large drifts in downstream connections. Furthermore, our setup balances positive and negative samples. However, the proportion of negative samples can be considerably larger in a real reconstruction scenario. As a consequence, we emphasize that these numbers must be considered together with the results reported in sections \ref{section:visualizing-graph} and \ref{section:easy-vs-hard}.

\begin{figure*}[ht!]
	\centering
    \includegraphics[width=\linewidth]{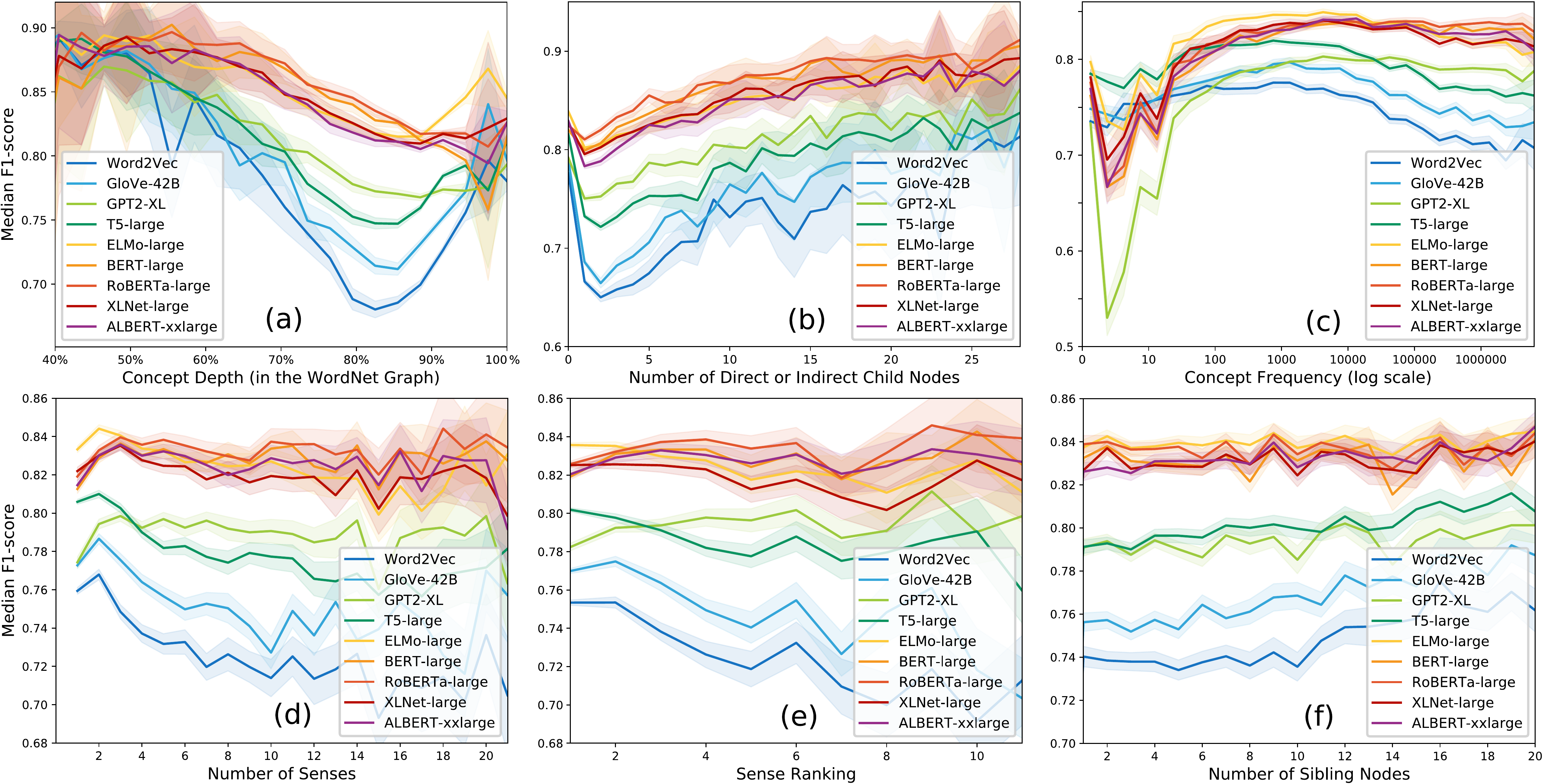}
    \caption{Semantic factors with a high (top charts) or low (bottom charts) impact on F1-score, along with their 90\% confidence intervals. Charts only display ranges where at least 100 samples existed.
    Appendix \ref{section:aditional-factors} shows additional factors along with the specific implementation details.}
	\label{fig:factors}
\end{figure*}


\subsection{Extracting the Knowledge Graph}
\label{section:visualizing-graph}

Predicting a knowledge graph has a complexity of at least $O(N^2)$ in the number of analyzed concepts. In our case, this imposes a highly demanding computational obstacle because WordNet has over $82000$ noun synsets. To accelerate experimentation and facilitate our analysis and visualizations, we focus on extracting a WordNet sub-graph comprising $46$ nodes not seen during training or validation. These nodes are picked to include easily recognizable relations. We use the tree-edit-distance to evaluate how close are the reconstructed graphs to the target graph extracted from WordNet. Table \ref{table:tree-edit-distances} shows our results. 



Table \ref{table:tree-edit-distances} shows that graphs retrieved using CE models are closer to the target than graphs provided by NCE and GLM models. In particular, the best results are achieved by BERT, ALBERT, and XLNet, indicating that these models encode more accurate semantic information than the alternative models. These results are consistent with those obtained in Section \ref{section:prediction-performance-comparison}. The graphs for all the models can be found in Appendix \ref{section:all-reconstructed-graphs}.

\section{What is easy or hard? What are these models learning?}
\label{section:easy-vs-hard}

Section \ref{section:how-accurate} shows that different model families differ in their errors. Furthermore, it shows that within the same family, models have similar biases. In this section, we elucidate which semantic factors impact the performance of these models and which ones do not affect their F1-score.

Figure \ref{fig:factors}-a shows that most models decrease their F1-score as concepts get more specific. We hypothesize that higher-level concepts (e.g., \textit{Animal}) appear more frequently and in more diverse contexts, as they are also seen as instances of their sub-classes (e.g., \textit{Dog}, \textit{Cat}, \textit{Chihuahua}), allowing the models to learn more precise representations for them. In contrast, lower-level concepts will only appear in specific contexts (e.g., texts about \textit{Apple-Head-Chihuahua}). Figure \ref{fig:factors}-b corroborates this intuition, as concepts with a higher number of sub-classes have higher F1-scores. Figure \ref{fig:factors}-c shows that models degrade their F1-score when concepts are too frequent. In particular, NCE and GLM models are more sensitive to this factor.

Another finding is that CE and GLM models are almost unaffected by the number of senses that a certain word has, neither to their sense ranking or their number of sibling concepts, displaying almost flat charts (see Figures \ref{fig:factors}-d-e-f). This result suggests that these models pay more attention to the context than to the target word. This behavior is opposed to what NCE models exhibit according to \citet{Yaghoobzadeh:2019}, as NCE models tend to focus more on frequent senses.





\begin{table*}[ht]
    \centering
    \begin{tabular}{l l c c c c c c C{0.8cm} c c}
\cline{3-11}
\textbf{}  &  \textbf{}  &  \multirow{1.7}{*}{\footnotesize{\textbf{artifact}}}  &
\multirow{1.7}{*}{\footnotesize{\textbf{attribute}}}  &
\footnotesize{\vspace*{-0.6mm}\textbf{living}}  &
\multirow{1.7}{*}{\footnotesize{\textbf{matter}}}  &
\multirow{1.7}{*}{\footnotesize{\textbf{person}}}  &
\multirow{1.7}{*}{\footnotesize{\textbf{relation}}} & 
\multirow{1.7}{*}{\footnotesize{\textbf{part}}} &
\footnotesize{\textbf{social}}  & 
\footnotesize{\textbf{taxonomic}} \\ [-0.4ex]
\footnotesize{\textbf{Family}}  &
\footnotesize{\textbf{Model}}  & 
&
& 
\footnotesize{\vspace*{0.2mm}\textbf{thing}}  &
&
&
&
&
\footnotesize{\vspace*{0.2mm}\textbf{group}}  & 
\footnotesize{\vspace*{0.2mm}\textbf{group}} \\ [-0.4ex]
\hline
\multirow{1.7}{*}{\footnotesize{NCE}} &  \footnotesize{Word2Vec}  &  \footnotesize{.7120}  &  \footnotesize{.7044}  &  \footnotesize{.7295}  &  \footnotesize{.7402}  &  \footnotesize{.7208}  &
\footnotesize{.7264}  &
\footnotesize{.7532}  & \footnotesize{.7497}  &  \footnotesize{.6920} \\ [-0.4ex]
 &  \footnotesize{GloVe-42B}  &  \footnotesize{.7389}  &  \footnotesize{.7213}  &  \footnotesize{.7421}  &  \footnotesize{.7633}  &  \footnotesize{.7351}  & 
 \footnotesize{.7567}  &
 \footnotesize{.7759} & \footnotesize{.7579}  &  \footnotesize{.6648}  \\ [-0.4ex]
\hline
\multirow{1.7}{*}{\footnotesize{GLM}} &  \footnotesize{GPT-2}  &  \footnotesize{.7903}  &  \footnotesize{.7730}  &  \footnotesize{.7300}  &  \footnotesize{.7582}  &  \footnotesize{.7207}  &
\footnotesize{.7612}  &
\footnotesize{.7540} &  \footnotesize{.8155}  &  \footnotesize{.3030}   \\ [-0.4ex]
 &  \footnotesize{T5}  &  \footnotesize{.7868}  &  \footnotesize{.7649}  &  \footnotesize{.7862}  &  \footnotesize{.8002}  &  \footnotesize{.7735}  &
 \footnotesize{.7963}  &
 \footnotesize{.8051}  &  \footnotesize{.7868}  &  \footnotesize{.6944} \\ [-0.4ex]
\hline
\multirow{4.4}{*}{\footnotesize{CE}} &  \footnotesize{ELMo}  &  \footnotesize{.8308}  &  \footnotesize{.8093}  &  \footnotesize{.8187}  &  \footnotesize{.7756}  &  \footnotesize{.8022}  &
\footnotesize{.7679}  &
\footnotesize{.7580}  &  \footnotesize{.8312}  &  \footnotesize{.6011} \\ [-0.4ex]
 &  \footnotesize{BERT}  &  \footnotesize{.8249}  &  \footnotesize{.8094}  &  \footnotesize{.7593}  &  \footnotesize{.7645}  &  \footnotesize{.7379}  &
 \footnotesize{.7662}  &
 \footnotesize{.7499}  &  \footnotesize{.8516}  &  \footnotesize{.4804} \\ [-0.4ex]
 &  \footnotesize{RoBERTa}  &  \footnotesize{.8315}  &  \footnotesize{.8167}  &  \footnotesize{.7823}  &  \footnotesize{.7614}  &  \footnotesize{.7585}  &
 \footnotesize{.7649}  &
 \footnotesize{.7441}  &  \footnotesize{.8552}  &  \footnotesize{.4921} \\ [-0.4ex]
 &  \footnotesize{XLNet}  &  \footnotesize{.8319}  &  \footnotesize{.8064}  &  \footnotesize{.7907}  &  \footnotesize{.7636}  &  \footnotesize{.7779}  &
 \footnotesize{.7659}  &
 \footnotesize{.7526}  &  \footnotesize{.8422}  &  \footnotesize{.5371} \\ [-0.4ex]
 &  \footnotesize{ALBERT}  &  \footnotesize{.8231}  &  \footnotesize{.8050}  &  \footnotesize{.7758}  &  \footnotesize{.7685}  &  \footnotesize{.7826}  &
 \footnotesize{.7727}  &
 \footnotesize{.7610}  &  \footnotesize{.8556}  &  \footnotesize{.4277} \\ [-0.4ex]
 \hline
    \end{tabular}
    \caption{Average F1-score for some semantic categories revealing models strengths and weaknesses. Several other categories are reported in Appendix \ref{section:categories-f1-full} along with their standard deviations.}
    \label{table:category-f1}
\end{table*}

\begin{figure*}[ht]
	\centering
    \includegraphics[width=\linewidth]{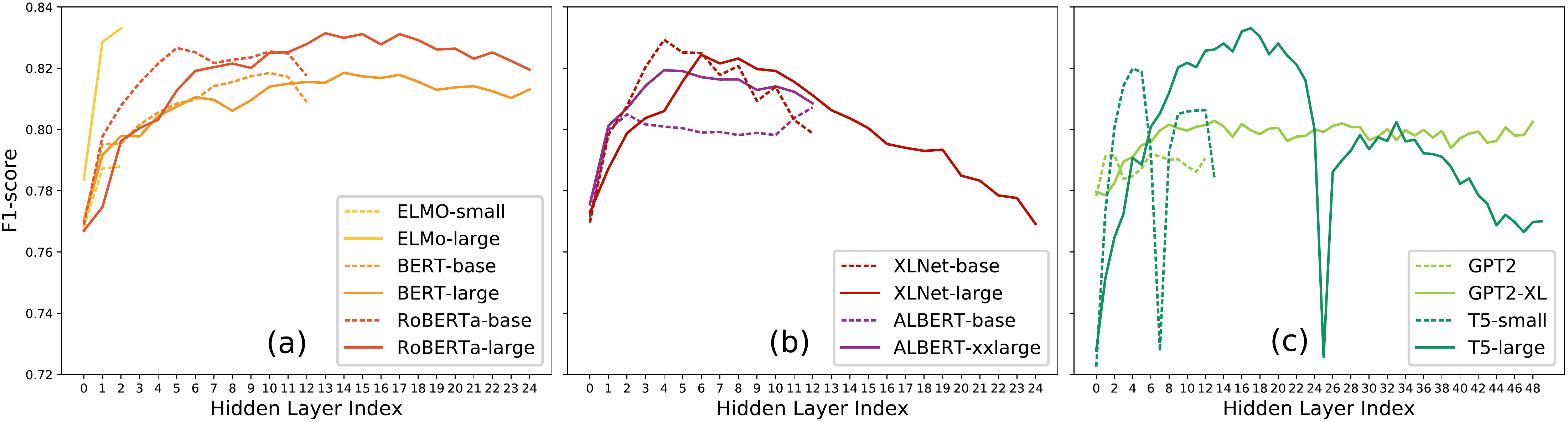}
    \caption{F1-score for hypernym prediction across each model layer.}
	\label{fig:layer-analysis}
\end{figure*}

In most cases, the same family models have similar behaviors, especially within the NCE or CE families. Also, different families show different patterns. Table \ref{table:category-f1} shows some salient examples. Surprisingly, all models struggle in the category ``taxonomic groups''. Manual inspection of sentences makes us believe that the context confuses CE and GLM models in these cases. In many sentences, the corresponding concept could be nicely replaced by another, conveying a modified but still valid message. This phenomenon does not occur in other categories such as ``social group'' or ``attribute'', even though these concepts are closely related to ``taxonomic groups''.

\section{Where is this knowledge located?}
\label{section:models-inner-workings}

As mentioned in Section \ref{section:related-work}, prior work has not shown consensus about where is semantic information encoded inside these architectures. Our experiments shed light on this subject. Figure \ref{fig:layer-analysis} shows how each layer contributes to the F1-score. 

Figures \ref{fig:layer-analysis}-a and \ref{fig:layer-analysis}-b show the performance across layers for the CE-based models. They reveal that while BERT and RoBERTa use their top-layers to encode semantic information, XLNet and ALBERT use the first layers. Figure \ref{fig:layer-analysis}-c shows that while GPT-2 uses all its layers to encode semantics, T5 shows an M shape related to its encoder-decoder architecture. The chart shows that T5 uses its encoder to hold most of the semantic information. We also note that small models show similar patterns as their larger counterparts.

\section{Further discussion and implications}
\label{section:discussion}

\begin{table*}[h]
    \centering
    \begin{tabular}{@{\hskip 0in}l@{\hskip 0.06in}p{8.4cm}@{\hskip 0.1in}p{1.9cm}@{\hskip 0.1in}p{2.6cm}@{\hskip 0in}}
&   &  \footnotesize{\textbf{Supporting}}  &  \\ [-0.2em]
& \footnotesize{\textbf{Finding}}  &  \footnotesize{\textbf{Evidence}}  &  \footnotesize{\textbf{Involved Models}} \\
\hline
\footnotesize{(1)} & \footnotesize{All models encode a relevant amount of knowledge about semantic relations in WordNet, but this knowledge contains imprecisions.  } &  \scriptsize{All}  &  \scriptsize{All}  \\
\hline
\footnotesize{(2)} & \footnotesize{The ability to learn concept relations depends on how frequent and specific the concepts are. Some model families are more affected.} &  \footnotesize{Fig. \ref{fig:factors}a-c}  & \scriptsize{NCE and GLM}  \\
\hline
\footnotesize{(3)} & \footnotesize{Concept difficulty is usually homogeneous within each model family. Some semantic categories challenge all models.} & \footnotesize{Table \ref{table:category-f1}}   & \scriptsize{All}  \\ 
\hline
\footnotesize{(4)} & \footnotesize{Some models encode stronger semantic knowledge than others, usually according to their family.} &  \footnotesize{Tables \ref{table:models-performance}, \ref{table:tree-edit-distances}, \ref{table:category-f1}}  &  \scriptsize{ELMo, BERT, RoBERTa, ALBERT, XLNet, T5 }  \\
\hline
\footnotesize{(5)} & \footnotesize{Some models focus their encoding of semantic knowledge in specific layers, and not distributed across all layers.} &  \footnotesize{Table \ref{table:models-performance}, Fig. \ref{fig:layer-analysis}}  & \scriptsize{GLM}  \\
\hline
\footnotesize{(6)} &\footnotesize{Models have distinctive patterns as to where they encode semantic knowledge. Patterns are model-specific and not size-specific.} &  \footnotesize{Table \ref{table:models-performance}}, Fig. \ref{fig:layer-analysis}  & \scriptsize{All}  \\
\hline
\footnotesize{(7)} & \footnotesize{Model size has an impact in the quality of the captured semantic knowledge, as seen in our layer-level probe tests.} &  \footnotesize{Table \ref{table:models-performance}}, Fig. \ref{fig:layer-analysis}  & \scriptsize{ELMo, RoBERTa, ALBERT, GPT-2, T5 }  \\
\hline
\footnotesize{(8)} & \footnotesize{Semantic knowledge does not depend on pre-training corpus size.} &  \footnotesize{Tables \ref{table:models-performance}, \ref{section:corpus-size-comparison}-\ref{table:corpus-sizes}}  & \scriptsize{-}  \\
\hline 
\footnotesize{(9)} & \footnotesize{Contextual models are unaffected by multi-sense words.} & \footnotesize{Fig. \ref{fig:factors}}d-f & \scriptsize{CE and GLM} \\
\hline
    \end{tabular}
    \caption{Summary of our main findings and their corresponding supporting evidence.}
    \label{table:findings}
\end{table*}

Table \ref{table:findings} summarizes our main findings. Findings (1), (2), and (3) indicate that, to a different extent, all models encode relevant knowledge about the hierarchical semantic relations included in WordNet. However, as we mention in Section \ref{section:easy-vs-hard}, we observe that the ability to learn about a concept depends on its frequency in the training corpus and the specificity of its meaning. Furthermore, some concept categories seem to be hard for every model family, while some are particularly difficult for contextual models such as CE. We hypothesize that stronger inductive biases are required to capture low-frequency concepts. Furthermore, we believe that new learning approaches are needed to discriminate accurate meaning for high-frequency concepts. As expected, our findings indicate that model families have different biases leading to different behaviors. Thus, our results can illuminate further research to improve semantic capabilities by combining each family of models' strengths. For example, one could combine them as ensembles, each one equipped with a different loss function (i.e., one generative approach resembling GLM-based methods and another discriminative resembling CE-based methods).

Findings (4), (5), and (6) suggest that instead of a standard finetuning of all layers of BERT according to a given downstream task, to improve semantic capabilities, one could perform a task profiling to decide the best architecture for the task and also how to take advantage of it. Using only a limited number of layers or choosing a different learning rate for each layer, one could exploit the semantic knowledge that the pre-trained model carries, avoiding the degradation of this information present at the top layers, especially when using T5, XLNet, or ALBERT-large. Accordingly, recent work on adaptive strategies to output predictions using a limited number of layers \citep{xin:2020:deebert, liu:2020:fastbert, hou:2020dynabert, Schwartz:2020, Fan:2020:Reducing, bapna:2020} would benefit from using architectures that encode knowledge in the first layers. To the best of our knowledge, these works have only used BERT and RoBERTa, achieving a good trade-off between accuracy and efficiency. Only \citet{zhou:2020bertpatience} has explored ALBERT, reporting improved accuracy by stopping earlier. Our findings explain this behavior and suggest that T5 or XLNet may boot their results even further as these architectures have sharper and higher information peaks in their first layers.

Findings (7) and (8) suggest that 
recent success in semantic NLP tasks might be due more to the use of larger models than large corpora for pretraining. This also suggests that to improve model performance in semantic tasks, one could train larger models even without increasing the corpus size. A similar claim has been proposed by \cite{Li:2020TrainLT} leading to empirical performance improvements. 

Finally, finding (9) is important because it suggests that contextual models pay as much attention to the context as to the target word and are probably biased in favor of contextual information, even if they are not based on the Masked-Language-Model strategy. We believe that this inductive bias could be exploited even further in the design of the underlying architecture. Thus this finding might elucidate a design direction to encourage more effective learning of semantic knowledge.


\section{Related work}
\label{section:related-work}



The success of deep learning architectures in various NLP tasks has fueled a growing interest to improve understanding of what these models encode. Studies like \citet{Tenney:19a} claim that success in a specific task helps understand what type of information the model encodes. \\ 

\textbf{Evidence of syntactic information}: Using probing classifiers, \citet{Clark:19} claims that some specific BERT's attention heads show correspondence with syntactic tasks. \citet{Goldberg:19} illustrates the capabilities that BERT has to solve syntactic tasks, such as subject-verb agreement. \citet{Hewitt:19} proposes a structural probe that evaluates whether syntax trees are encoded in a linear transformation of BERT embeddings. The study provides evidence that syntax trees are implicitly embedded in BERT's vector geometry. \citet{Reif:19} has found evidence of syntactic representation in BERT's attention matrices, with specific directions in space representing particular dependency relations. \\

\textbf{Evidence of semantic information}: \citet{Reif:19} suggests that BERT's internal geometry may be broken into multiple linear subspaces, with separate spaces for different syntactic and semantic information. Despite this result, previous work has not yet reached a consensus about this topic. While some studies show satisfactory results in tasks such as entity types \citep{Tenney:19b}, semantic roles \citep{Rogers:20}, and sentence completion \citep{Ettinger:20}, other studies show less favorable results in coreference \citep{Tenney:19a}, Multiple-Choice Reading Comprehension \citep{Si:19} and Lexical Relation Inference \cite{Levy:15}, claiming that BERT's performance may not reflect the model's true ability of language understanding and reasoning.  \citet{Tenney:19a} proposes a set of edge probing tasks to test the encoded sentential structure of contextualized word embeddings. The study shows evidence that the improvements that BERT and GPT-2 offer over non contextualized embeddings as GloVe is only significant in syntactic-level tasks. Regarding static word embeddings, \citet{Yaghoobzadeh:2019} shows that senses are well represented in single-vector embeddings if they are frequent and that this does not harm NLP tasks whose performance depends on frequent senses.

\textbf{Layer-wise or head-wise information}: \citet{Tenney:19b} shows that the first layers of BERT focus on encoding short dependency relationships at the syntactic level (e.g., subject-verb agreement). In contrast, top layers focus on encoding long-range dependencies (e.g., subject-object dependencies). \citet{Peters:2018} 
supports similar declarations for Convolutional, LSTM, and self-attention architectures.
While these studies also support that the top layers appear to encode semantic information, the evidence to support this claim is not conclusive or contradictory with other works.
For example,
\citet{Jawahar:19} could only identify one SentEval semantic task that topped at the last layer. In terms of information flow, \citet{Voita:19b} reports that information about the past in left-to-right language models gets vanished as the information flows from bottom to top BERT's layers. \citet{Hao:19} shows that the lower layers of BERT change less during finetuning, suggesting that layers close to inputs learn more transferable language representations. \citet{Press:20} shows that increasing self-attention at the bottom layers improves language modeling performance based on BERT. Other studies focus on understanding how self-attention heads contribute to solving specific tasks \citep{Vig:19}. \citet{Kovaleva:19} shows a set of attention patterns repeated across different heads when trying to solve GLUE tasks~\citep{Wang:18}. Furthermore, \citet{Michel:19} and \citet{Voita:19a} show that several heads can be removed without harming downstream tasks.

\textbf{Automated extraction of concept relations}: Although the main focus of our work is not to master the probing task of extracting knowledge from WordNet, but to use it as an instrument to verify and compare the abilities of current families of language models to encode this kind of knowledge, for completitude we include a brief mention of previous literature regarding this subject.
Relation extraction is an active research topic. Early works are either feature-based, usually relying on SVMs, Maximum Entropy, or on a set of manually defined rules \cite{Hearst:98, Kambhatla:04, Dashtipour:07, Minard:11, Weeds:14, Chen:15}. Other methods rely on manually defined distance metrics to estimate the relatedness of two semantic instances \cite{Danan:12, Panyam:16}.
Following works have used different types of neural networks or LSTM modules for this task \cite{Liu:13, Zeng:14, Zeng:15, Zhang:15, Song:18}, or attention-based and transformer-based mechanisms with outstanding results \cite{Zhou:16, Soares:19, Huang:20, Qin:21, Zhong:21}. \\

\textbf{Alternative approaches}: Several alternative approaches have been used in previous works. Some are dataset-focused \cite{Miller:94, Levy:15, Wang:18, Wiedemann:19}, usually relying on annotated corpora that challenge semantic abilities. These approaches have provided useful insights, but usually suffer from low availability of data as they usually cover a small fraction of the WordNet ontology. As an example, BLESS \cite{Baroni:11} includes gold-standard annotations for only 200 concepts.
Other approaches have tested semantic ability by using prompt-engineering and inspecting the predictions of the models \cite{Petroni:19, Ettinger:20, Talmor:20}, but other works have also shown a high variability in the results depending on the prompt design \cite{Balasubramanian:20, Reynolds:21, Zhao:2021}.

\section{Conclusions}
\label{section:conclusions}
In this work, we exploit the semantic conceptual taxonomy behind WordNet to test the ability of current families of pre-trained language models to learn semantic knowledge from massive sources of unlabeled data. Our main conclusion is that, indeed, to a significant extent, these models learn relevant knowledge about the organization of concepts in WordNet, but also contain several imprecisions. 
We also notice that different families of models present dissimilar behavior, suggesting the encoding of different biases. 

We hope our study helps to inspire new ideas to improve the semantic learning abilities of current pre-trained language models.  

\section*{Acknowledgments}
We thank the anonymous reviewers, whose helpful comments led to increased clarity and improved analysis. We would also like to thank all the members of the IALab research group for their helpful feedback and suggestions. We acknowledge Alex Medina for his help in the early stages of this project.
This work was partially funded by the Millennium Institute for Foundational Research on Data and the Fondecyt project number 1200211.

\bibliographystyle{acl_natbib}
\bibliography{acl2021}

\clearpage

\appendix

\section{Implementation details}
\label{app:imp}

\subsection{Edge probing classifier details}
\label{app:imp-1}

To study the extent to which these Language Models deal with semantic knowledge, we extend the methodology introduced by \citet{Tenney:19a}. In that study, the authors defined a probing classifier at the sentence level, training a supervised classifier with a task-specific label. The probing classifier's motivation consists of verifying when the sentence's encoding help to solve a specific task, quantifying these results for different word embeddings models. We cast this methodology to deal with semantic knowledge extracted from WordNet. Rather than working at the sentence level, we define an edge probing classifier that learns to identify if two concepts are semantically related.

To create the probing classifier, we retrieve all the glosses from the Princeton WordNet Gloss Corpus. The dataset provides WordNet's synsets gloss with manually matched words identifying the context-appropriate sense. 

As a reference of size, the selected annotations in the corpus accounted for $41502$ lemmas, corresponding to $34371$ WordNet synsets. This resulted in $230215$ valid WordNet relations.

\begin{figure*}[h!]
    \centering
    \includegraphics[width=9cm]{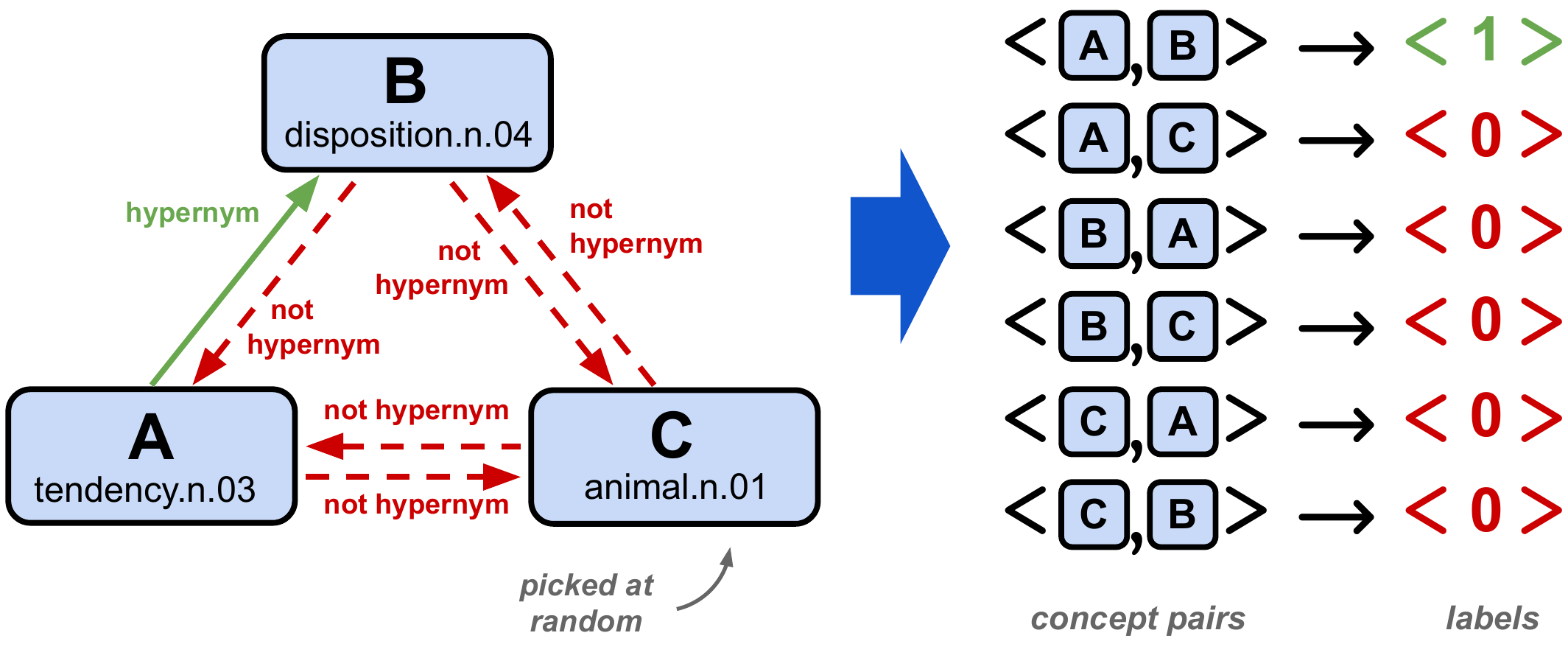}
    \caption{Each triplet is used to create related and unrelated pairs of words according to the relationship \textit{hypernym of}. We create six edge probing pairs, and therefore, the edge probing classifier will need to identify the pair's components and the order in which the words were declared in the pair.}
    \label{fig:expl2}
\end{figure*}

In WordNet, each sense is coded as one of the synsets related to the concept (e.g., sense \textit{tendency.n.03} for the word tendency). Using a synset A and its specific sense provided by the tagged gloss, we retrieve from WordNet one of its direct or indirect hypernyms, denoted as B (see Figure \ref{fig:expl2}). If WordNet defines two or more hypernyms for A, we choose one of them at random. We sample a third synset C, at random from an unrelated section of the taxonomy, taking care that C is not related to either A or B (e.g., \textit{animal.n.01}). Then, $\langle A, B, C \rangle$ form a triplet that allows us to create six testing edges for our classifier: $\langle A, B \rangle$, which is compounded by a pair of related words through the semantic relation \textit{hypernym of}, and five pairs of unrelated words ($\langle A, C \rangle$, $\langle B, C \rangle$, $\langle B, A \rangle$, $\langle C, A \rangle$, $\langle C, B \rangle$). We associate a label to each of these pairs that show whether the pair is related or not (see Figure \ref{fig:expl2}). Note that we define directed edges, meaning that the pair $\langle A, B \rangle$ is related, but $\langle B, A \rangle$ is unrelated to the relationship \textit{hypernym of}. Accordingly, the edge probing classifier will need to identify the pair's components and the order in which the concepts were declared in the pair.

We create training and testing partitions ensuring that each partition has the same proportion of leaves versus internal nodes. The latter is essential to identify related pairs. During training, we guarantee that each training synset is seen at least once by the probing classifier. To guarantee the above, we sample each synset in the training set and sample some of its hypernyms at random. Then. we randomly sample some unrelated synset for each related pair that has no relation to any of the words in the related pair. We create three partitions from this data on 70/15/15 for training, development, and testing foldings, respectively.

We train the MLP classifier using a weighted binary cross-entropy loss function. Since we have one positive and five negative examples per triplet, we use a weighted loss function with weights 5 and 1 for the positive and negative class, respectively. Accordingly, positive and negative examples have the same relevance during training. We implemented the linear layer and the MLP classifier using a feed forward network with 384 hidden units. The MLP was trained using dropout at 0.425 and a $L_2$ regularizer to avoid overfitting. 


To create the vector representations for each of the word embeddings models considered in this study, we concatenate the hidden state vectors of all the layers for each tagged synset. For both CE and GLM-based models, each gloss was used as a context to build specific contextual word embeddings. If the gloss has more than one tagged token, we take only the first of them for the analysis.

\subsection{WordNet metrics: distance}
\label{section:graph-distance-methodoogy}

Lets say that we name \textit{``Case-1"} if $y$ is ancestor of $x$, and \textit{``Case-2"} otherwise. Let $d_W(x,y)$ be the Wordnet distance between two synsets $x$, $y$, defined by:
\begin{equation}
    d_W(x, y) = \begin{cases}
               d_{\text{path}}(x, y)    & \text{\textit{\footnotesize{Case-1}}}, \\
               d_{\text{path}}(x, z) + d_{\text{path}}(y, z)    & \text{\textit{\footnotesize{Case-2}}}, 
           \end{cases}
\end{equation}

\noindent where $d_{\text{path}}(x, y)$ is the length of the shortest path between $x$ and $y$ in WordNet, measured in number of hops, and $z$ is the closest common ancestor of $x$ and $y$ in the case that $y$ is not an ancestor of $x$. 

\subsection{Minimum-Spanning-Arborescence optimization problem}
\label{msa-formulation}

Given a graph $G$ with nodes $N$ and unknown edges $E$, we define an auxiliary graph $G'$ with nodes $N$ and edges $E'$, comprised of all possible directed edges. For each edge $e \in E'$, we obtain a prediction $h_e$ that estimates the probability of that edge representing a valid hypernymy relation, and a distance $d_e$ that estimates the ``parent closeness''\footnote{The value of this distance will be small if the hypernym relation is close, or large if it is distant or not valid.} between the nodes in $G$.

We define $\delta(v)$ to be the set of edges $\{\langle u,v \rangle: u \in N, u \neq v\}$ where edge $\langle u,v \rangle$ represents a $\langle \text{parent},\text{child} \rangle$ relation. We also define $\gamma(S)$ to be the set of edges $\{\langle u,v \rangle \in E': u \notin S, v \in S\}$. We estimate the graph topology of $G$ defined by $E \subset E'$ by solving the following optimization problem:
\begin{equation} \max_{r \in N} \quad {\sum_{e \in E'}{x_e h_e}} \quad  \text{ s.t.} \quad  x_e \in X^* \label{eq:best-root} \end{equation} 
\begin{equation} X^* = \argmin {\sum_{e \in E'}{x_e d_e}} \label{eq:minimum-spanning-arborescence}
\end{equation}
\begin{equation} \text{s.t.} \!\begin{cases}
\hspace*{1.0em} x_e \in \{0,1\} \quad & e \in E' \\
\hspace*{1.0em} \sum_{e \in \delta(v)}{x_e} = 1 \quad & \forall v \in N \setminus \{r\}  \\
\hspace*{1.0em} \sum_{e \in \gamma(S)}{x_e} \geq 1 \quad & \forall S \subset N \setminus \{r\} 
\end{cases} \label{eq:minimum-spanning-arborescence}
\end{equation}

Objective function (\ref{eq:best-root}) is used to find the best root node $r$; and the nested optimization problem (\ref{eq:minimum-spanning-arborescence}) is the minimum spanning arborescence problem applied to the dense graph $G'$. The final binary values of $x_e$ estimate $E$ by indicating if every possible edge $e$ exist in the graph or not. To solve this optimization problem, we need estimates of $h_e$ and $d_e$ for each edge $e$. We use the output of the probing classifier as an estimate of the probability of $h_e$, and use TIM and MCM scores as estimates for $d_e$ (See Section \ref{reconstruction-methodology}).

\section{Pre-Training corpus comparison} \label{section:corpus-size-comparison}

\begin{table}[H]
\caption{Pre-Training corpus sizes used for each one of the studied models. The official sources report corpus sizes in terms of number of tokens or uncompressed size in GB. The symbol * denotes values estimated by us based on official available information. Sizes represents uncompressed corpus sizes.}
\label{table:corpus-sizes}
\begin{center}
\begin{tabular}{llcc}
    \hline
    \multicolumn{1}{c}{\multirow{2}{*}{\textbf{Family}}} & \multicolumn{1}{c}{\multirow{2}{*}{\textbf{Model}}} & \multicolumn{2}{c}{\textbf{Corpus Size}} \\ \cline{3-4}
    {} & {} & \textbf{Tokens} & \textbf{Size} \\
    \hline
    \multirow{2}{*}{NCE} & Word2Vec  & 33B & 150GB* \\
                         & GloVe-42B & 42B & 175GB* \\ \hline
    \multirow{2}{*}{GLM} & GPT-2     & 10B* & 40GB \\
                         & T5        & 180B* & 750GB \\ \hline
    \multirow{5}{*}{CE}  & ELMo      & 0.8B & 4GB* \\
                         & BERT      & 3.9B & 16GB \\
                         & RoBERTa   & 38.7B* & 160GB \\
                         & XLNet     & 32.9B & 140GB* \\
                         & ALBERT    & 3.9B & 16GB \\ \hline
\end{tabular}
\end{center}
\end{table}

\section{Additional Reconstructed Graphs\footnote{Due to space restrictions, the graphs corresponding to Word2Vec, ELMo, T5, BERT will only be included in an extended version of this paper, uploaded to ArXiv}} \label{section:all-reconstructed-graphs}

\begin{figure}[H]
    \centering
    \includegraphics[width=110pt]{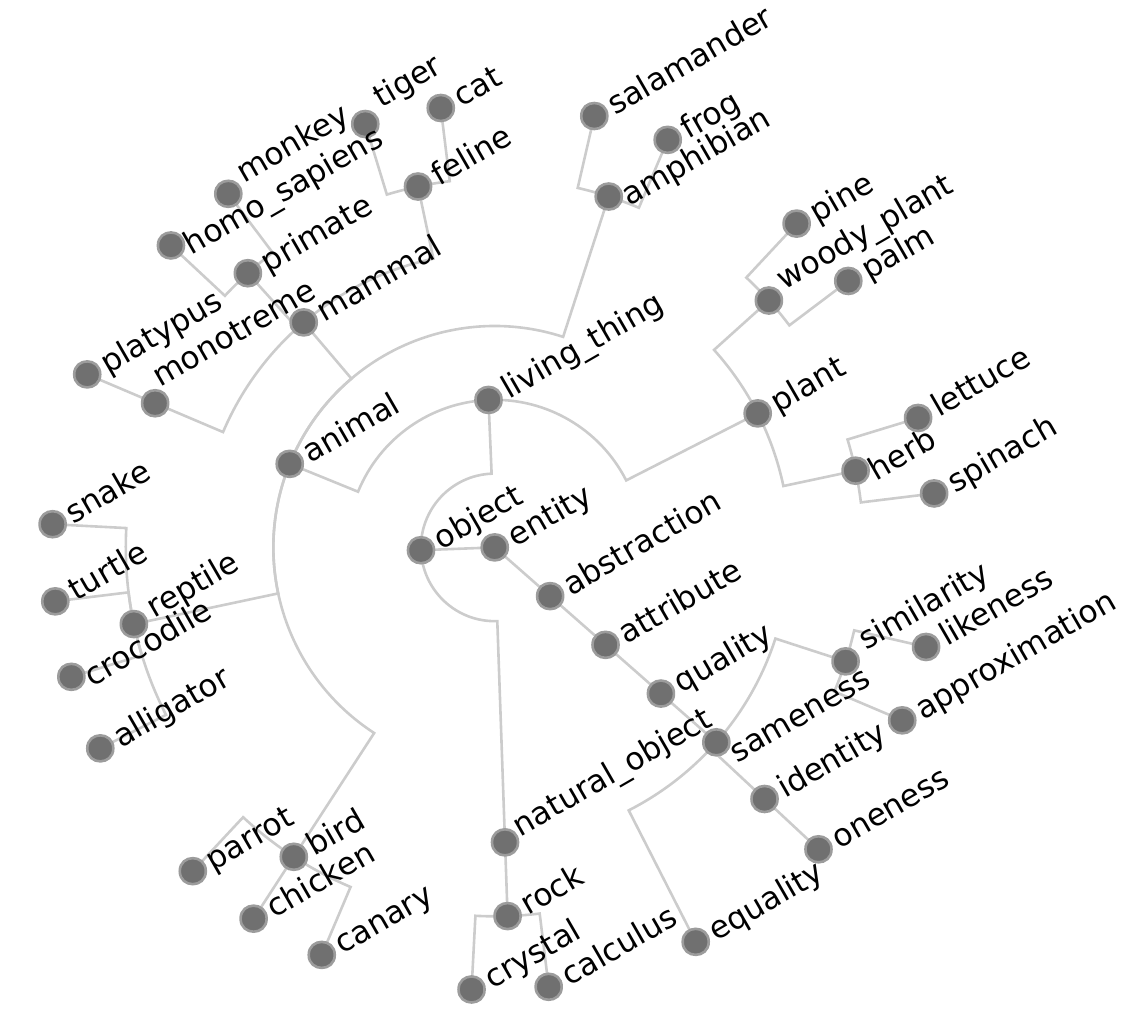}
    \caption{Ground Truth Knowledge Graph}
\end{figure}

\begin{figure}[H]
    \centering
	\includegraphics[width=100pt]{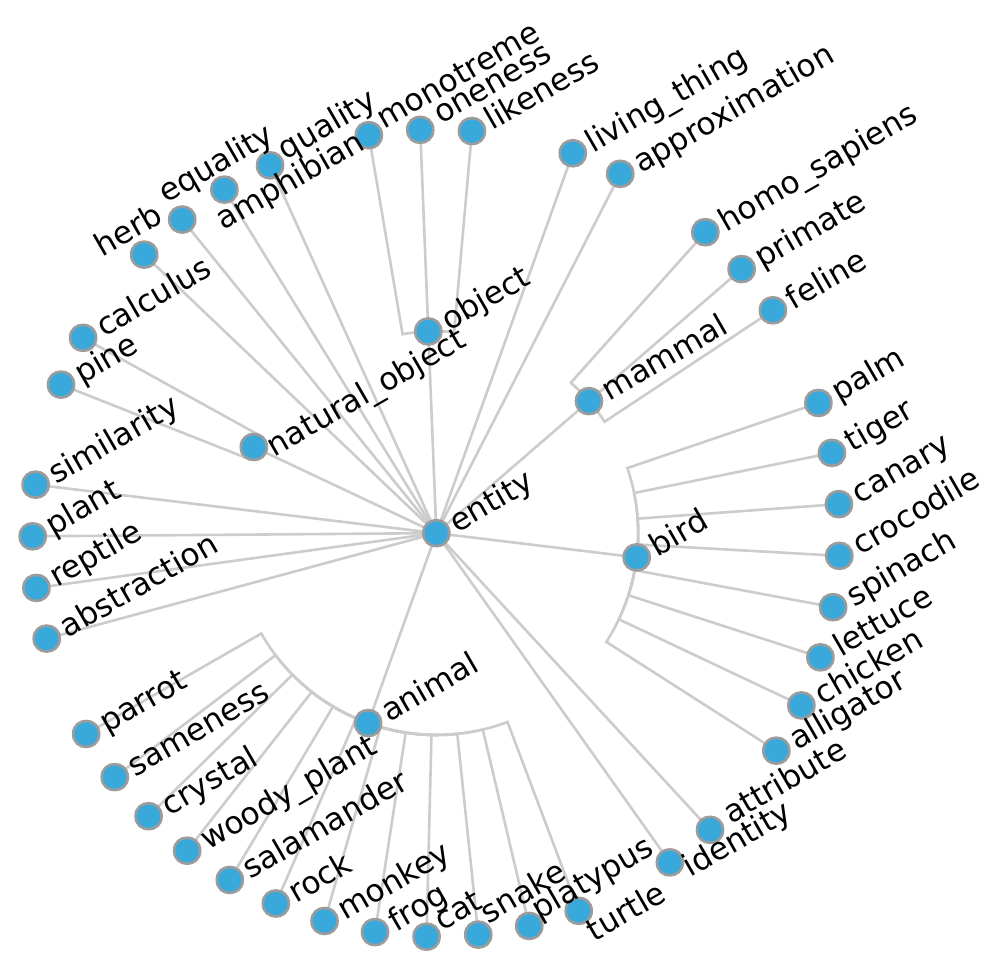}
	\caption{GloVe-42B reconstruction using TIM}
\end{figure}

\begin{figure}[H]
		\includegraphics[width=100pt]{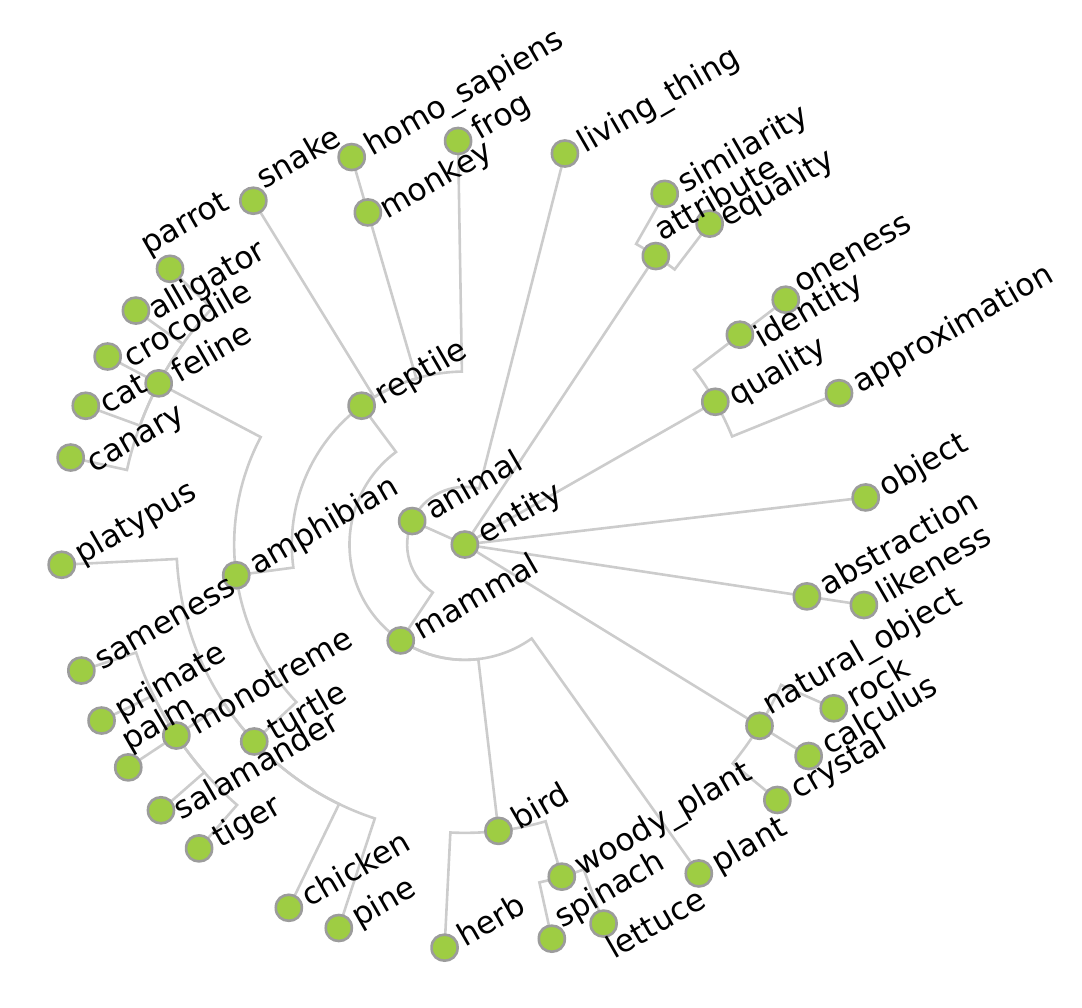}
		\caption{GPT-2-XL reconstruction using TIM}
\end{figure}

\begin{figure}[H]
		\includegraphics[width=100pt]{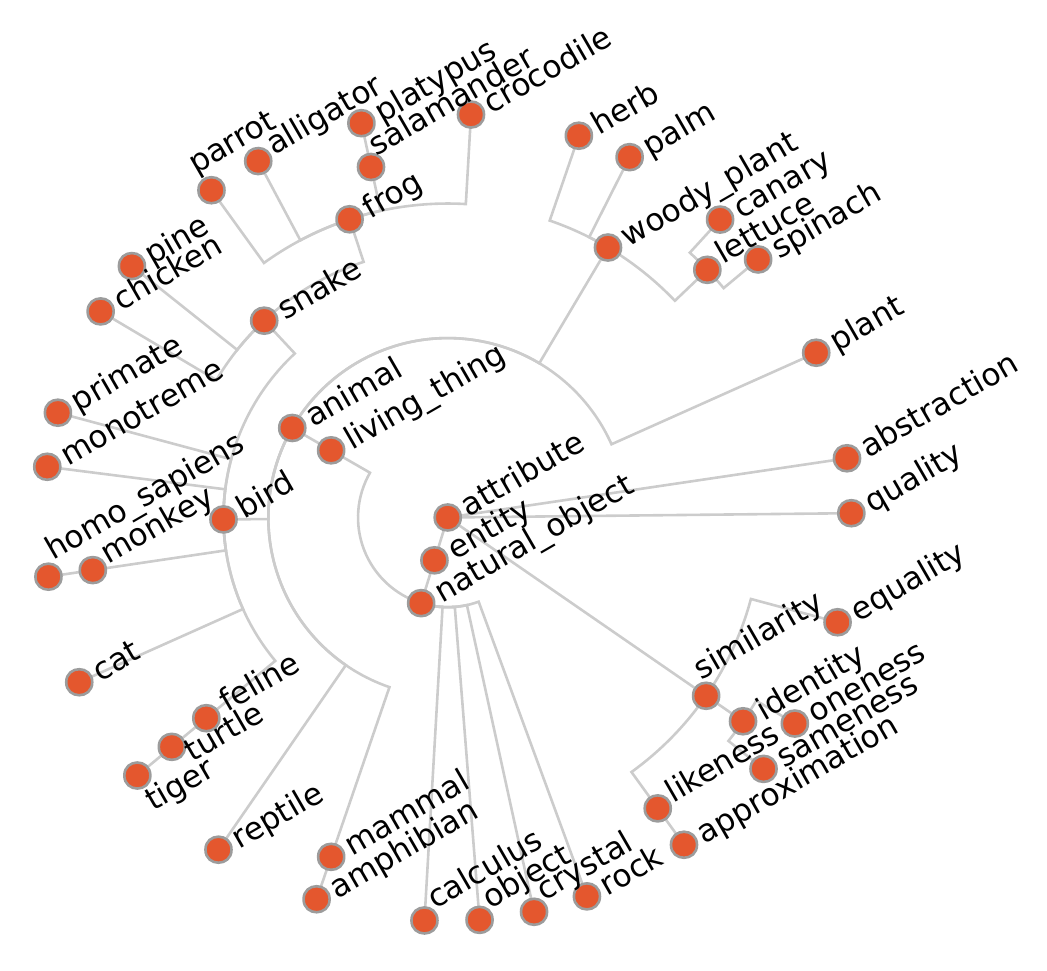}
		\caption{RoBERTa-large reconstruction using TIM}
\end{figure}

\begin{figure}[H]
		\includegraphics[width=100pt]{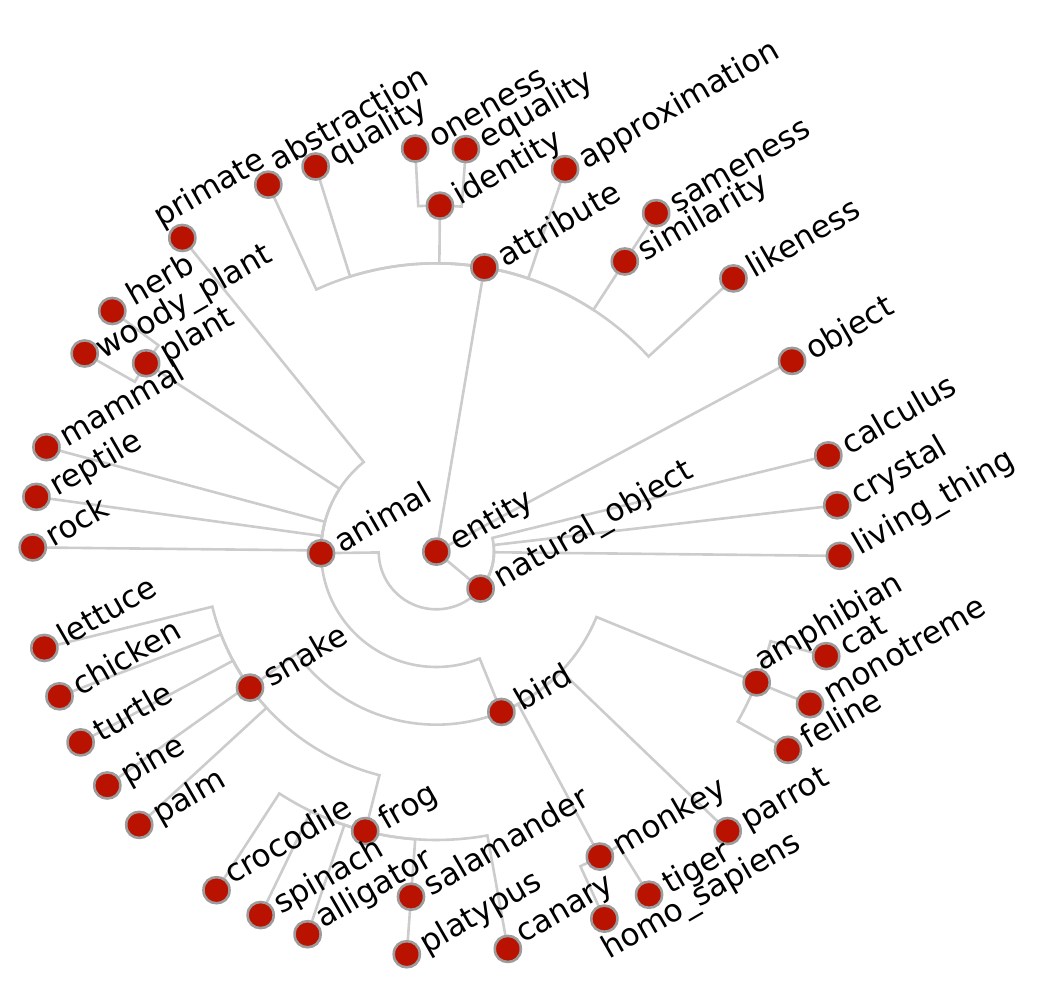}
		\caption{XLNet-large reconstruction using TIM}
\end{figure}
\begin{figure}[H]
		\includegraphics[width=100pt]{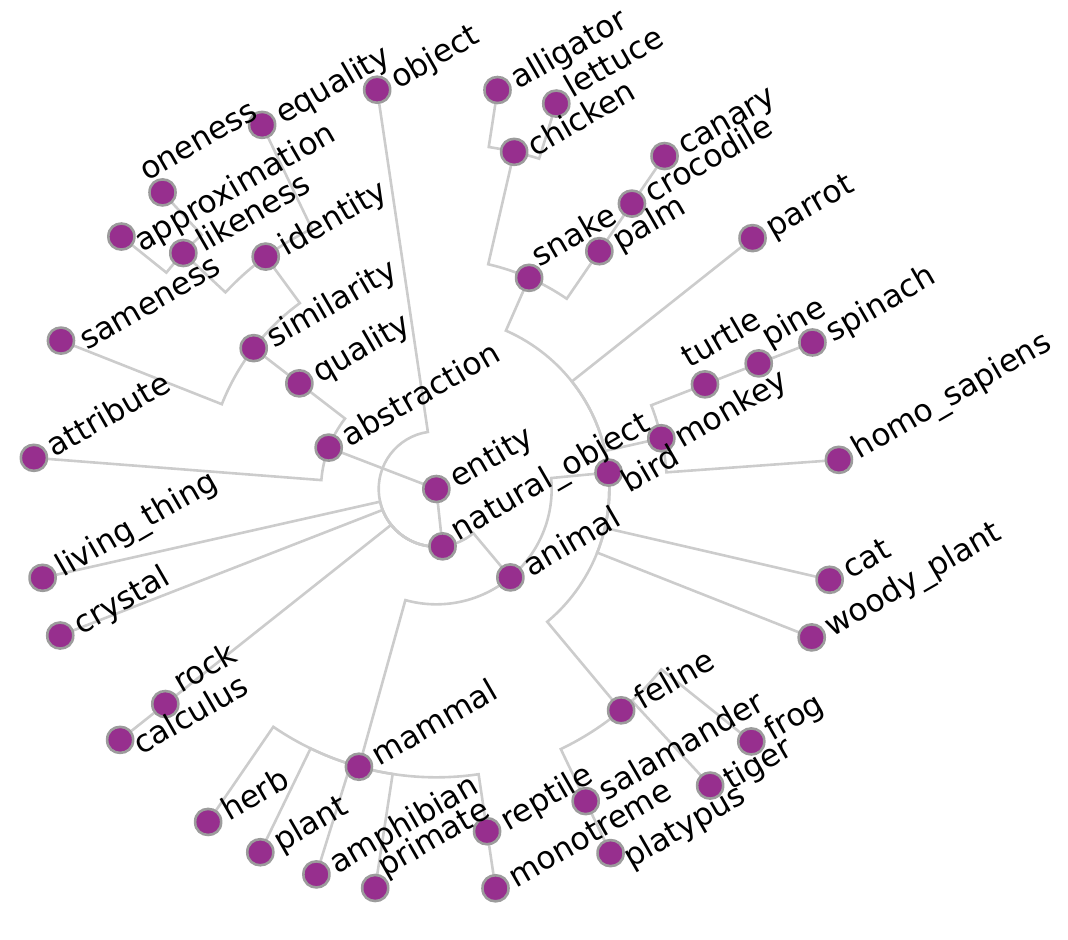}
		\caption{ALBERT-large reconstruction using TIM}
\end{figure}


\section{Further information about the impact of semantic factors.\footnote{Due to space restrictions, other factors and graphs will only be included in an extended version of this paper, uploaded to ArXiv}} \label{section:aditional-factors}

\textbf{Relative depth in the WordNet graph}: (Figure \ref{fig:factors}-a). For each synset, we compared F1 with depth score (0 \% for the root and 100 \% for leaves) measuring differences between higher/lower level concepts. 

\textbf{Concept frequency}: In Figure \ref{fig:factors}-c we evaluate if frequent concepts are easier or harder to capture for these models. The frequency was computed by counting occurrences in the 38 GB of OpenWebText Corpus (\url{http://Skylion007.github.io/OpenWebTextCorpus}).

\textbf{Number of Senses and Sense Ranking}: (Figure \ref{fig:factors}-d-e) We studied if models are impacted by multi-sense concepts such as ``period'', and by their sense ranking (how frequent or rare those senses are). Surprisingly contextualized models, and specially CE models have no significant impact by this factor, suggesting that these models are very effective at deducing the correct sense based on their context. These charts also suggest that these models may be considering context even more than the words themselves. This is intuitive for Masked-Language-Models such as BERT, but not for others, such as GPT-2. Non-contextualized models are impacted by this factor, as expected.

\begin{figure}[H]
    \centering
    \includegraphics[height=130pt]{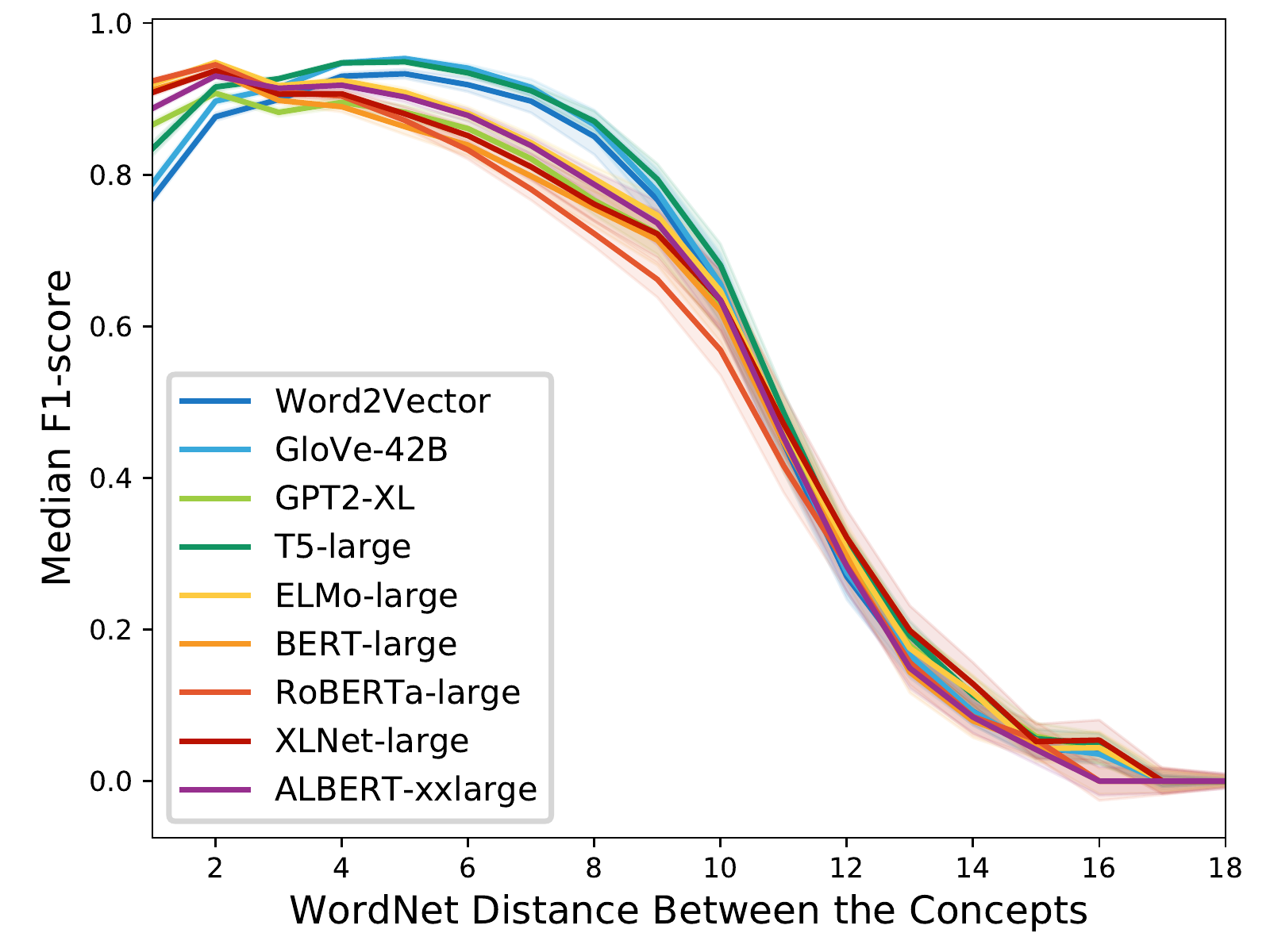}
	\caption{\textbf{Graph distance between concepts}: We measured the impact of the number of ``hops'' that separate two tested concepts on pair-wise F1 score. This chart reveals a strong correlation of all the models in this aspect. As an example of this phenomenon, closer relations such as $\langle \text{chihuahua},\text{dog} \rangle$ are, in general, considerably easier to capture than distant relations such as $\langle \text{chihuahua},\text{entity} \rangle$. For details on how we implement the distance in WordNet, check Appendix \ref{section:graph-distance-methodoogy}.}
	\label{fig:factors2-distances}
\end{figure}

\newpage

\clearpage
\onecolumn
\section{F1-scores of additional categories}
\label{section:categories-f1-full}
\begin{table*}[hbt!]
    \centering
    \begin{tabular}{l|c@{\hskip 0.06in}c|c@{\hskip 0.1in}c|c@{\hskip 0.1in}c@{\hskip 0.05in}c@{\hskip 0.06in}c@{\hskip 0.06in}c}
Category  &  \small{W2V}  &  \small{GloVe}  &  \small{GPT-2}  &  \small{T5}  &  \small{ELMo}  &  \small{BERT}  &  \small{RoBERTa}  &  \small{XLNet}  &  \small{ALBERT} \\ \hline {} & & & & & & & & \\ [-1.9ex]
\multirow{2}{*}{\footnotesize{abstraction}}  &  \small{.7142}  &  \small{.7296}  &  \small{.7224}  &  \small{.7662}  &  \small{.7808}  &  \small{.7718}  &  \small{.7759}  &  \small{.7712}  &  \small{.7635} \\ [-1.3ex]
  &  \scriptsize{${\pm}$ .1277}  &  \scriptsize{${\pm}$ .1194}  &  \scriptsize{${\pm}$ .1897}  &  \scriptsize{${\pm}$ .0942}  &  \scriptsize{${\pm}$ .1203}  &  \scriptsize{${\pm}$ .1582}  &  \scriptsize{${\pm}$ .1537}  &  \scriptsize{${\pm}$ .1404}  &  \scriptsize{${\pm}$ .1732} \\ [+0.8ex]
\multirow{2}{*}{\footnotesize{{} {} {} {} attribute}}  &  \small{.7044}  &  \small{.7213}  &  \small{.7730}  &  \small{.7649}  &  \small{.8093}  &  \small{.8094}  &  \small{.8167}  &  \small{.8064}  &  \small{.8050} \\ [-1.3ex]
  &  \scriptsize{${\pm}$ .1310}  &  \scriptsize{${\pm}$ .1237}  &  \scriptsize{${\pm}$ .0911}  &  \scriptsize{${\pm}$ .0863}  &  \scriptsize{${\pm}$ .0886}  &  \scriptsize{${\pm}$ .0998}  &  \scriptsize{${\pm}$ .0926}  &  \scriptsize{${\pm}$ .0891}  &  \scriptsize{${\pm}$ .0998} \\ [+0.8ex]
\multirow{2}{*}{\footnotesize{{} {} {} {} communication}}  &  \small{.6974}  &  \small{.7251}  &  \small{.7826}  &  \small{.7587}  &  \small{.8049}  &  \small{.8246}  &  \small{.8249}  &  \small{.8066}  &  \small{.8093} \\ [-1.3ex]
  &  \scriptsize{${\pm}$ .1330}  &  \scriptsize{${\pm}$ .1224}  &  \scriptsize{${\pm}$ .0967}  &  \scriptsize{${\pm}$ .1083}  &  \scriptsize{${\pm}$ .0925}  &  \scriptsize{${\pm}$ .0979}  &  \scriptsize{${\pm}$ .0983}  &  \scriptsize{${\pm}$ .0987}  &  \scriptsize{${\pm}$ .1023} \\ [+0.8ex]
\multirow{2}{*}{\footnotesize{{} {} {} {} group}}  &  \small{.7068}  &  \small{.6929}  &  \small{.4972}  &  \small{.7262}  &  \small{.6821}  &  \small{.6173}  &  \small{.6256}  &  \small{.6491}  &  \small{.5858} \\ [-1.3ex]
  &  \scriptsize{${\pm}$ .1320}  &  \scriptsize{${\pm}$ .1339}  &  \scriptsize{${\pm}$ .2955}  &  \scriptsize{${\pm}$ .1179}  &  \scriptsize{${\pm}$ .1711}  &  \scriptsize{${\pm}$ .2399}  &  \scriptsize{${\pm}$ .2305}  &  \scriptsize{${\pm}$ .2139}  &  \scriptsize{${\pm}$ .2745} \\ [+0.8ex]
\multirow{2}{*}{\footnotesize{{} {} {} {} {} {} {} {} social group}}  &  \small{.7497}  &  \small{.7579}  &  \small{.8155}  &  \small{.7868}  &  \small{.8312}  &  \small{.8516}  &  \small{.8552}  &  \small{.8422}  &  \small{.8556} \\ [-1.3ex]
  &  \scriptsize{${\pm}$ .1058}  &  \scriptsize{${\pm}$ .1046}  &  \scriptsize{${\pm}$ .0883}  &  \scriptsize{${\pm}$ .0867}  &  \scriptsize{${\pm}$ .0707}  &  \scriptsize{${\pm}$ .0742}  &  \scriptsize{${\pm}$ .0698}  &  \scriptsize{${\pm}$ .0724}  &  \scriptsize{${\pm}$ .0819} \\ [+0.8ex]
\multirow{2}{*}{\footnotesize{{} {} {} {} {} {} {} {} taxonomic group}}  &  \small{.6920}  &  \small{.6648}  &  \small{.3030}  &  \small{.6944}  &  \small{.6011}  &  \small{.4804}  &  \small{.4921}  &  \small{.5371}  &  \small{.4277} \\ [-1.3ex]
  &  \scriptsize{${\pm}$ .1306}  &  \scriptsize{${\pm}$ .1330}  &  \scriptsize{${\pm}$ .2025}  &  \scriptsize{${\pm}$ .1208}  &  \scriptsize{${\pm}$ .1583}  &  \scriptsize{${\pm}$ .2025}  &  \scriptsize{${\pm}$ .1903}  &  \scriptsize{${\pm}$ .1944}  &  \scriptsize{${\pm}$ .2305} \\ [+0.8ex]
\multirow{2}{*}{\footnotesize{{} {} {} {} {} {} {} {} {} {} {} {} {} family}}  &  \small{.7412}  &  \small{.7213}  &  \small{.3131}  &  \small{.6691}  &  \small{.5461}  &  \small{.5626}  &  \small{.5379}  &  \small{.5733}  &  \small{.5437} \\ [-1.3ex]
  &  \scriptsize{${\pm}$ .1363}  &  \scriptsize{${\pm}$ .1244}  &  \scriptsize{${\pm}$ .2003}  &  \scriptsize{${\pm}$ .1276}  &  \scriptsize{${\pm}$ .1630}  &  \scriptsize{${\pm}$ .1537}  &  \scriptsize{${\pm}$ .1502}  &  \scriptsize{${\pm}$ .1626}  &  \scriptsize{${\pm}$ .1705} \\ [+0.8ex]
\multirow{2}{*}{\footnotesize{{} {} {} {} {} {} {} {} {} {} {} {} {} genus}}  &  \small{.6267}  &  \small{.6040}  &  \small{.2567}  &  \small{.7156}  &  \small{.6167}  &  \small{.3696}  &  \small{.4201}  &  \small{.4555}  &  \small{.2862} \\ [-1.3ex]
  &  \scriptsize{${\pm}$ .0989}  &  \scriptsize{${\pm}$ .1127}  &  \scriptsize{${\pm}$ .1582}  &  \scriptsize{${\pm}$ .1001}  &  \scriptsize{${\pm}$ .1301}  &  \scriptsize{${\pm}$ .1857}  &  \scriptsize{${\pm}$ .1855}  &  \scriptsize{${\pm}$ .1853}  &  \scriptsize{${\pm}$ .1945} \\ [+0.8ex]
\multirow{2}{*}{\footnotesize{{} {} {} {} psychological feature}}  &  \small{.7256}  &  \small{.7478}  &  \small{.7829}  &  \small{.7795}  &  \small{.8163}  &  \small{.8181}  &  \small{.8229}  &  \small{.8077}  &  \small{.8208} \\ [-1.3ex]
\small{{} {} {} {} }  &  \scriptsize{${\pm}$ .1122}  &  \scriptsize{${\pm}$ .1016}  &  \scriptsize{${\pm}$ .0904}  &  \scriptsize{${\pm}$ .0778}  &  \scriptsize{${\pm}$ .0851}  &  \scriptsize{${\pm}$ .0954}  &  \scriptsize{${\pm}$ .0930}  &  \scriptsize{${\pm}$ .0931}  &  \scriptsize{${\pm}$ .0915} \\ [+1.3ex]
\multirow{2}{*}{\footnotesize{{} {} {} {} relation}}  &  \small{.7264}  &  \small{.7567}  &  \small{.7612}  &  \small{.7963}  &  \small{.7679}  &  \small{.7662}  &  \small{.7649}  &  \small{.7659}  &  \small{.7727} \\ [-1.3ex]
  &  \scriptsize{${\pm}$ .1304}  &  \scriptsize{${\pm}$ .1042}  &  \scriptsize{${\pm}$ .0809}  &  \scriptsize{${\pm}$ .0688}  &  \scriptsize{${\pm}$ .0878}  &  \scriptsize{${\pm}$ .0995}  &  \scriptsize{${\pm}$ .0982}  &  \scriptsize{${\pm}$ .0908}  &  \scriptsize{${\pm}$ .0929} \\ [+0.8ex]
\hline {} & & & & & & & & \\ [-1.9ex]
\multirow{2}{*}{\footnotesize{artifact}}  &  \small{.7120}  &  \small{.7389}  &  \small{.7903}  &  \small{.7868}  &  \small{.8308}  &  \small{.8249}  &  \small{.8315}  &  \small{.8319}  &  \small{.8231} \\ [-1.3ex]
  &  \scriptsize{${\pm}$ .1194}  &  \scriptsize{${\pm}$ .1068}  &  \scriptsize{${\pm}$ .0736}  &  \scriptsize{${\pm}$ .0676}  &  \scriptsize{${\pm}$ .0693}  &  \scriptsize{${\pm}$ .0742}  &  \scriptsize{${\pm}$ .0700}  &  \scriptsize{${\pm}$ .0702}  &  \scriptsize{${\pm}$ .0761} \\ [+0.8ex]
\multirow{2}{*}{\footnotesize{{} {} {} {} covering}}  &  \small{.7230}  &  \small{.7510}  &  \small{.7903}  &  \small{.7878}  &  \small{.8398}  &  \small{.8392}  &  \small{.8363}  &  \small{.8393}  &  \small{.8322} \\ [-1.3ex]
  &  \scriptsize{${\pm}$ .1097}  &  \scriptsize{${\pm}$ .0970}  &  \scriptsize{${\pm}$ .0713}  &  \scriptsize{${\pm}$ .0606}  &  \scriptsize{${\pm}$ .0599}  &  \scriptsize{${\pm}$ .0706}  &  \scriptsize{${\pm}$ .0571}  &  \scriptsize{${\pm}$ .0576}  &  \scriptsize{${\pm}$ .0621} \\ [+0.8ex]
\multirow{2}{*}{\footnotesize{{} {} {} {} instrumentality}}  &  \small{.7064}  &  \small{.7378}  &  \small{.7930}  &  \small{.7902}  &  \small{.8308}  &  \small{.8134}  &  \small{.8337}  &  \small{.8313}  &  \small{.8233} \\ [-1.3ex]
  &  \scriptsize{${\pm}$ .1219}  &  \scriptsize{${\pm}$ .1052}  &  \scriptsize{${\pm}$ .0728}  &  \scriptsize{${\pm}$ .0648}  &  \scriptsize{${\pm}$ .0676}  &  \scriptsize{${\pm}$ .0748}  &  \scriptsize{${\pm}$ .0691}  &  \scriptsize{${\pm}$ .0711}  &  \scriptsize{${\pm}$ .0763} \\ [+0.8ex]
\multirow{2}{*}{\footnotesize{{} {} {} {} {} {} {} {} {} device}}  &  \small{.7097}  &  \small{.7435}  &  \small{.7956}  &  \small{.7899}  &  \small{.8311}  &  \small{.8198}  &  \small{.8358}  &  \small{.8326}  &  \small{.8230} \\ [-1.3ex]
  &  \scriptsize{${\pm}$ .1200}  &  \scriptsize{${\pm}$ .1009}  &  \scriptsize{${\pm}$ .0713}  &  \scriptsize{${\pm}$ .0667}  &  \scriptsize{${\pm}$ .0689}  &  \scriptsize{${\pm}$ .0701}  &  \scriptsize{${\pm}$ .0640}  &  \scriptsize{${\pm}$ .0675}  &  \scriptsize{${\pm}$ .0743} \\ [+0.8ex]
\hline {} & & & & & & & & \\ [-1.9ex]
\multirow{2}{*}{\footnotesize{causal agent}}  &  \small{.7240}  &  \small{.7398}  &  \small{.7253}  &  \small{.7751}  &  \small{.8022}  &  \small{.7453}  &  \small{.7631}  &  \small{.7788}  &  \small{.7826} \\ [-1.3ex]
  &  \scriptsize{${\pm}$ .1137}  &  \scriptsize{${\pm}$ .1101}  &  \scriptsize{${\pm}$ .1105}  &  \scriptsize{${\pm}$ .0757}  &  \scriptsize{${\pm}$ .0884}  &  \scriptsize{${\pm}$ .1093}  &  \scriptsize{${\pm}$ .1056}  &  \scriptsize{${\pm}$ .1035}  &  \scriptsize{${\pm}$ .0934} \\ [+0.8ex]
\multirow{2}{*}{\footnotesize{{} {} {} {} person}}  &  \small{.7208}  &  \small{.7351}  &  \small{.7207}  &  \small{.7735}  &  \small{.8022}  &  \small{.7379}  &  \small{.7585}  &  \small{.7779}  &  \small{.7826} \\ [-1.3ex]
  &  \scriptsize{${\pm}$ .1135}  &  \scriptsize{${\pm}$ .1113}  &  \scriptsize{${\pm}$ .1132}  &  \scriptsize{${\pm}$ .0746}  &  \scriptsize{${\pm}$ .0854}  &  \scriptsize{${\pm}$ .1101}  &  \scriptsize{${\pm}$ .1054}  &  \scriptsize{${\pm}$ .1053}  &  \scriptsize{${\pm}$ .0937} \\ [+0.8ex]
\hline {} & & & & & & & & \\ [-1.9ex]
\multirow{2}{*}{\footnotesize{living thing}}  &  \small{.7295}  &  \small{.7421}  &  \small{.7300}  &  \small{.7862}  &  \small{.8187}  &  \small{.7593}  &  \small{.7823}  &  \small{.7907}  &  \small{.7758} \\ [-1.3ex]
  &  \scriptsize{${\pm}$ .1112}  &  \scriptsize{${\pm}$ .1078}  &  \scriptsize{${\pm}$ .1004}  &  \scriptsize{${\pm}$ .0749}  &  \scriptsize{${\pm}$ .0850}  &  \scriptsize{${\pm}$ .0997}  &  \scriptsize{${\pm}$ .0922}  &  \scriptsize{${\pm}$ .0928}  &  \scriptsize{${\pm}$ .0909} \\ [+0.8ex]
\multirow{2}{*}{\footnotesize{{} {} {} {} animal}}  &  \small{.7349}  &  \small{.7391}  &  \small{.7515}  &  \small{.7837}  &  \small{.8389}  &  \small{.7914}  &  \small{.8179}  &  \small{.8135}  &  \small{.7781} \\ [-1.3ex]
  &  \scriptsize{${\pm}$ .1049}  &  \scriptsize{${\pm}$ .1028}  &  \scriptsize{${\pm}$ .0829}  &  \scriptsize{${\pm}$ .0714}  &  \scriptsize{${\pm}$ .0785}  &  \scriptsize{${\pm}$ .0801}  &  \scriptsize{${\pm}$ .0701}  &  \scriptsize{${\pm}$ .0737}  &  \scriptsize{${\pm}$ .0857} \\ [+0.8ex]
\multirow{2}{*}{\footnotesize{{} {} {} {} plant}}  &  \small{.7445}  &  \small{.7608}  &  \small{.7288}  &  \small{.8168}  &  \small{.8404}  &  \small{.7679}  &  \small{.7962}  &  \small{.7986}  &  \small{.7645} \\ [-1.3ex]
  &  \scriptsize{${\pm}$ .1044}  &  \scriptsize{${\pm}$ .0989}  &  \scriptsize{${\pm}$ .0842}  &  \scriptsize{${\pm}$ .0653}  &  \scriptsize{${\pm}$ .0692}  &  \scriptsize{${\pm}$ .0830}  &  \scriptsize{${\pm}$ .0688}  &  \scriptsize{${\pm}$ .0746}  &  \scriptsize{${\pm}$ .0861} \\ [+0.8ex]
\hline {} & & & & & & & & \\ [-1.9ex]
\multirow{1.5}{*}{\footnotesize{matter}}  &  \small{.7402}  &  \small{.7633}  &  \small{.7582}  &  \small{.8002}  &  \small{.7756}  &  \small{.7645}  &  \small{.7614}  &  \small{.7636}  &  \small{.7685} \\ [-1.3ex]
  &  \scriptsize{${\pm}$ .1117}  &  \scriptsize{${\pm}$ .1009}  &  \scriptsize{${\pm}$ .0834}  &  \scriptsize{${\pm}$ .0662}  &  \scriptsize{${\pm}$ .0886}  &  \scriptsize{${\pm}$ .0959}  &  \scriptsize{${\pm}$ .0941}  &  \scriptsize{${\pm}$ .0908}  &  \scriptsize{${\pm}$ .0938} \\ [+0.8ex]
\hline {} & & & & & & & & \\ [-1.9ex]
\multirow{1.5}{*}{\footnotesize{part}}  &  \small{.7532}  &  \small{.7759}  &  \small{.7540}  &  \small{.8051}  &  \small{.7580}  &  \small{.7499}  &  \small{.7441}  &  \small{.7526}  &  \small{.7610} \\ [-1.3ex]
  &  \scriptsize{${\pm}$ .1000}  &  \scriptsize{${\pm}$ .0852}  &  \scriptsize{${\pm}$ .0772}  &  \scriptsize{${\pm}$ .0595}  &  \scriptsize{${\pm}$ .0844}  &  \scriptsize{${\pm}$ .0977}  &  \scriptsize{${\pm}$ .0947}  &  \scriptsize{${\pm}$ .0880}  &  \scriptsize{${\pm}$ .0907} \\ [+0.8ex]
\hline {} & & & & & & & & \\ [-1.9ex]
\multirow{1.5}{*}{\footnotesize{substance}}  &  \small{.7560}  &  \small{.7791}  &  \small{.7508}  &  \small{.8073}  &  \small{.7542}  &  \small{.7436}  &  \small{.7370}  &  \small{.7477}  &  \small{.7580} \\ [-1.3ex]
  &  \scriptsize{${\pm}$ .0967}  &  \scriptsize{${\pm}$ .0809}  &  \scriptsize{${\pm}$ .0755}  &  \scriptsize{${\pm}$ .0567}  &  \scriptsize{${\pm}$ .0828}  &  \scriptsize{${\pm}$ .0952}  &  \scriptsize{${\pm}$ .0919}  &  \scriptsize{${\pm}$ .0862}  &  \scriptsize{${\pm}$ .0889} \\ [-1.9ex]
{} & & & & & & & &
    \end{tabular}
    \caption{Each value represents the mean F1-score and standard deviation of all the concepts that belong to each analyzed category. Only the larger version of each model is reported. This is not an extensive list and categories are somewhat imbalanced. Categories were selected based on the number of sub-categories they contained.}
    \label{tab:all-concepts-f1}
\end{table*}

\end{document}